\documentclass[10pt,twocolumn,letterpaper]{article}
\usepackage{titlesec}

\usepackage[pagenumbers]{iccv}      % To produce the REVIEW version
\usepackage{tikz}
\usepackage{tcolorbox}
\usepackage{appendix}
\usepackage{multirow}
\definecolor{iccvblue}{rgb}{0.21,0.49,0.74}
\usepackage[pagebackref,breaklinks,colorlinks,allcolors=iccvblue]{hyperref}
\usepackage[font=footnotesize,skip=0pt]{caption}
\titlespacing*{\paragraph}{10pt}{.5\baselineskip}{.5\baselineskip}

\newcommand{\ours}{\texttt{IntroStyle}\@\xspace}
\newcommand{\ourd}{\texttt{ArtSplit}\@\xspace}

\definecolor{tabfirst}{rgb}{1, 0.7, 0.7}
\definecolor{tabsecond}{rgb}{1, 0.85, 0.7}
\definecolor{tabthird}{rgb}{1, 1, 0.7}
\definecolor{correct}{rgb}{0.196, 0.705, 0.196}
\definecolor{wrong}{rgb}{0.705, 0.098, 0.098}
\definecolor{gold}{rgb}{0.843, 0.607, 0}
\definecolor{blue}{rgb}{0.423, 0.556, 0.749}
\definecolor{purple}{rgb}{0.604, 0, 1}

\def\paperID{55} % *** Enter the Paper ID here
\def\confName{ICCV}
\def\confYear{2025}

\title{\ours: Training-Free Introspective Style Attribution \\ using Diffusion Features}

\author{Anand Kumar, \quad Jiteng Mu, \quad Nuno Vasconcelos\\
University of California, San Diego\\
{\tt\small \{ank029, jmu, nvasconcelos\}@ucsd.edu}
}

\begin{document}
\maketitle
\begin{abstract}
Text-to-image (T2I) models have recently gained widespread adoption. This has spurred concerns about safeguarding intellectual property rights  and an increasing demand for mechanisms that prevent the generation of specific artistic styles. Existing methods for style extraction typically necessitate the collection of custom datasets and the training of specialized models. This, however, is resource-intensive, time-consuming, and often impractical for real-time applications. We present a novel, training-free framework to solve the style attribution problem, using the features produced by a diffusion model alone, without any external modules or retraining. This is denoted as {\it Introspective Style attribution\/} (\ours) and is shown to have superior performance to state-of-the-art models for style attribution. We also introduce a synthetic dataset of {\it Artistic Style Split\/} (\ourd) to isolate artistic style and evaluate fine-grained style attribution performance. Our experimental results on WikiArt and DomainNet datasets show that \ours is robust to the dynamic nature of artistic styles, outperforming existing methods by a wide margin. Website: \url{https://anandk27.github.io/IntroStyle}
\end{abstract}    
\section{Introduction}
\label{sec:intro}
\begin{figure}[t]
  \centering
   \includegraphics[width=1\columnwidth]{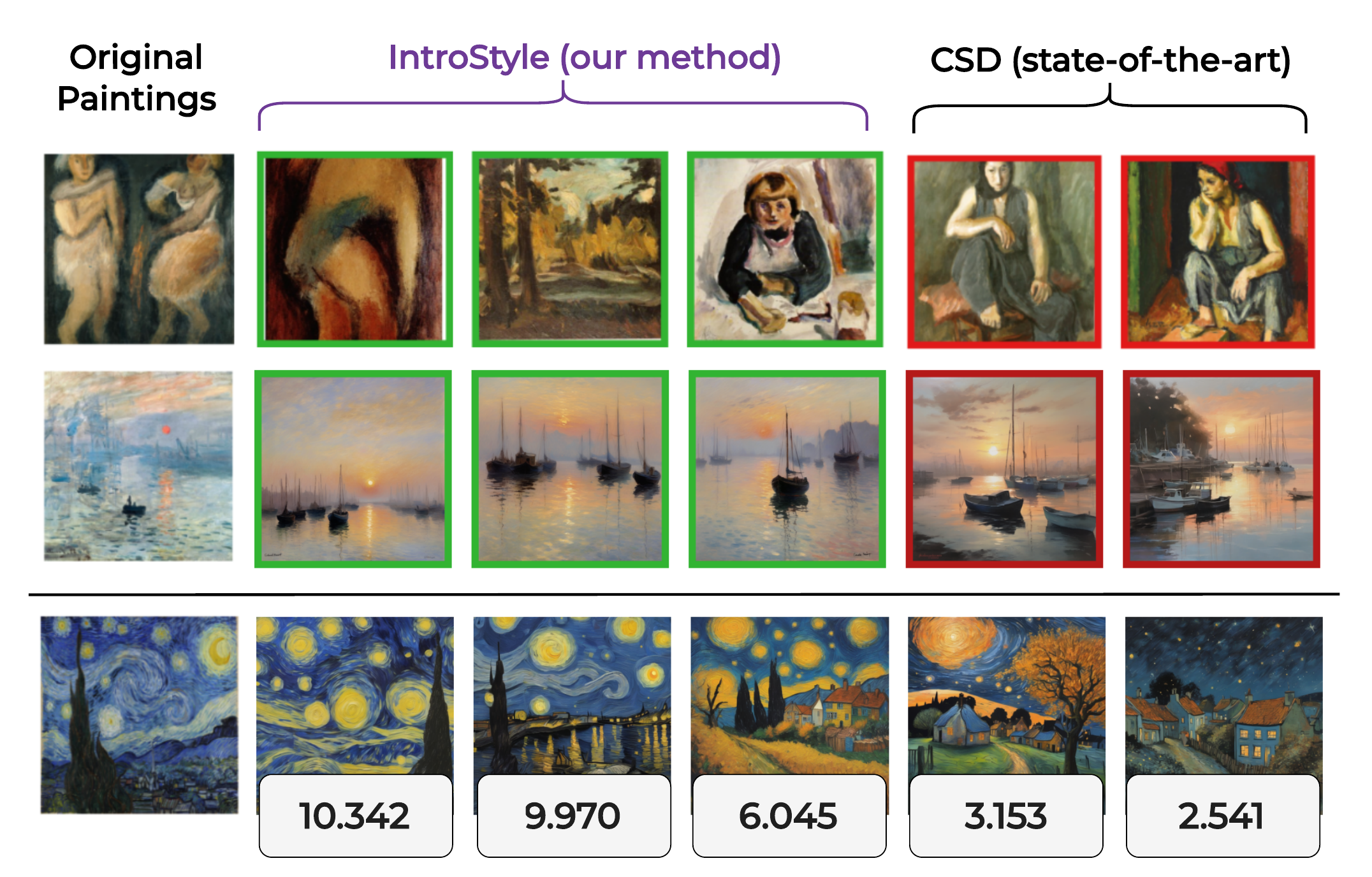}
   \caption{Introspective Style Attribution (\ours).  Top two rows: comparison of style attribution by \ours and a SOTA method (CSD).  A query image, the top-3 retrieval results of \ours, and the top-2 retrieval results of CSD are shown from left to right. Green colors indicate \textcolor{correct}{correct} and red \textcolor{wrong}{incorrect} retrievals. First row: WikiArt~\cite{saleh2015large} dataset. Second row: proposed synthetic Artistic Style Split (\ourd) dataset. While the SOTA style attributions are biased by image semantics, failing to retrieve images on the style of the query, \ours retrieves images of the correct style. Bottom: retrieval scores of \ours for \ourd synthetic images generated to have the semantic of the query (``Starry Night") but different styles, demonstrating the effectiveness of \ours as a metric for style measurement.}
   \label{fig:img_res}
\end{figure}
Diffusion models have significantly advanced image synthesis by applying iterative denoising processes guided by input prompts. Models like Stable Diffusion~\cite{rombach2022high}, DALL-E 3~\cite{betker2023improving}, and Imagen~\cite{saharia2022photorealistic} have emerged as a powerful paradigm for T2I synthesis, demonstrating remarkable capabilities in generating high-quality images from textual descriptions.   The success of diffusion-based approaches has led to their application in various domains, including layout-to-image generation, text-guided image generation, and even video synthesis. However, the widespread adoption and impressive performance of T2I modelss have also raised concerns about problems like copyright infringement. For effective performance, the models require large-scale pre-training on diverse datasets~\cite{schuhmann2022laion}. Since these datasets are collected automatically and dominantly from the web, it is difficult to control image provenance and avoid the collection of copyrighted imagery. This problem is compounded by the tendency of diffusion models to replicate elements from their training data~\cite{carlini2023extracting}. The legal implications of training models on copyrighted images have become a subject of recent debate and litigation, with artists arguing that the unauthorized use of their works for AI training~\cite{dryhurst2023ai, smh2022aifuture, growcoot2022midjourney} constitutes copyright infringement~\cite{lanz2023rutkowski, reuters2023aicopyright}.

Current mitigation approaches include ``unlearning" techniques to remove specific styles from AI models~\cite{wang2024data}, though these require costly model retraining and may not fully address indirect style replication through alternative prompts~\cite{petsiuk2025concept}. Style ``cloaking" methods~\cite{shan2023glaze} offer artists some protection but compromise the authentic viewing experience and place the burden of protection on creators. An alternative solution is to rely on attribution techniques~\cite{wang2023attribution, somepalli2024measuring, wang2024data}, which can be used to attribute synthesized images to different artistic styles, allowing the post-hoc assessment of how much they might, for example, replicate an artist's artwork. These methods are also more practical since they simply return examples from the dataset used to train the model, ranked by similarity to the synthesized image. The main difficulty is to develop similarity functions that focus on style and are not confused by other image properties, such as similar image content or semantics.

To solve this problem, prior attribution approaches have proposed retraining the diffusion model~\cite{wang2024data} or adding external models to evaluate data similarity~\cite{wang2023attribution, somepalli2024measuring}. This can add significant complexity since effective similarity models, like CLIP~\cite{radford2021learning} or DINO~\cite{caron2021emerging}, can be comparable in size to the diffusion model. On the other hand, prior diffusion model research has shown that different diffusion model layers develop distinct representations of higher and lower level image properties, such as structure vs color~\cite{voynov2023P+}, producing features that can be leveraged to solve problems like image correspondence~\cite{tang2023emergent} or segmentation~\cite{tian2024diffuse} without retraining. This led us to hypothesize that it may be possible to solve the attribution problem using the features produced by a diffusion model alone, without any external modules or retraining. We refer to this as {\it Introspective Style attribution\/} (\ours). 

In this work, we show that the hypothesis holds. We propose a simple and effective approach to differentiate image styles by analyzing the similarity of the features produced by a diffusion model for different images. We frame the denoising network of the diffusion model as an autoencoder whose encoder produces stylistic features. We then propose a metric to compare these features. While quite simple, this is shown to be quite effective for style retrieval. In fact, a comparison with the current state-of-the-art on popular style retrieval datasets shows that \ours significantly outperforms prior models trained in the existing evaluation datasets, as illustrated in the top rows of Fig.~\ref{fig:img_res}. Although fine-tuned for style similarity, existing models still tend to retrieve images of similar semantics, whereas \ours focuses much more uniquely on style. 

Beyond models, the problem of style attribution suffers from a shortage of evaluation datasets. Creating such datasets is far from trivial since even the concept of ``style" in art remains debatable. Nevertheless, many recognized artistic styles, such as surrealism and impressionism, are often associated with specific artists or movements. We use this social construct to define style as the collective global characteristics of an image identifiable with a particular artist or artistic movement. These characteristics encompass various elements such as color usage, brushstroke techniques, composition, and perspective. Under this definition, a single artist can produce art in several styles.  Current datasets, primarily focusing on comparisons between authentic artworks, provide limited ability for evaluating how strongly style attribution methods disentangles style from semantics. From this point of view, an ideal dataset for evaluating style attributions should contain images with all pairs of styles and semantics, a requirement only attainable by synthetic data. While CSD~\cite{somepalli2024measuring} introduces synthetic data for generated image evaluation and GDA~\cite{wang2023attribution} uses custom diffusion to generate training datasets, these datasets lack fine-grained separation of artistic style and image semantics, which are heavily correlated in their images. This makes the resulting models poorly suited for enforcing copyright protections~\cite{lin2024parrot}.
% these existing datasets fail to account for prompt manipulation techniques that can circumvent copyright protections \cite{tsai2023ring, chin2023prompting4debugging, wen2024hard}.

To address these problems, we introduce the {\it Artistic Style Split\/} (\ourd) dataset, inspired by~\cite{somepalli2024measuring}. This dataset combines real artistic image queries with synthetic retrieval images generated via diffusion models using carefully crafted prompts. For each authentic painting, we create two distinct prompts: a ``semantic" prompt that excludes stylistic elements and a ``style" prompt that omits semantic content. These prompts are fed into a diffusion model to generate corresponding images, enabling comprehensive coverage of the style $\times$ semantics space encompassing all combinations in real artistic queries. By analyzing the similarity between features extracted from query and synthetic images, we can precisely evaluate how feature representations balance style attribution against semantic similarity. Hence, style retrieval experiments on this dataset provide clear insights into how different models disentangle style from semantics. 
% This dataset is based on real artistic image queries and synthetic retrieval images generated by a diffusion model in response to prompts that combine stylistic and semantic descriptions. For each real painting, we produce a ``semantic" prompt, which does not mention style, and a ``style" prompt, which does not mention semantics. These prompts are then fed to a diffusion model to create corresponding images. This allows creating a dataset with very fined-grained coverage of the space of style $\times$ semantics, covering all possible combinations of these attributes that appear in real artistic queries. By measuring similarities between features extracted from the query and synthetic images, it is possible to perform fine-grained evaluations of how much the feature representation favors style attribution vs semantic similarity. Hence, style retrieval experiments on this dataset provide a clear picture of how different models overcome the problem of semantic bias. 
This is illustrated in the last row of Fig.~\ref{fig:img_res}, where the painting {\it Starry Night\/} by Vincent Van Gogh is used to retrieve images of paintings of the {\it Starry Night\/} semantic under various styles, using \ours. Note how images of style closer to the reference have higher similarity scores, providing evidence that the \ours representation favors style over semantics. This can be leveraged to implement style-based rejection sampling, preventing diffusion models from generating images of a specific style.

% The second row of Fig. \ref{fig:img_res} shows retrieval results on the {\it Style Hacks\/} (\ourd) dataset, where the proposed model outperforms a state-of-the-art approach. 

Overall, this paper makes the following contributions:
\begin{enumerate}
\item We formulate the problem of introspective style attribution without model retraining or external modules.
\item We propose \ours, a method to generate style descriptors from images and a metric to compare image pairs according to style.
\item We introduce the \ourd dataset to isolate artistic style and evaluate fine-grained style attribution performance.
\item We perform extensive experiments showing that \ours outperforms existing approaches of much higher complexity.
\end{enumerate}
% {\color{red} WHAT IS THE TEASER?}

\section{Related Work}

\subsection{Style Transfer}
%Early computer vision approaches attempted to model style using low-level visual features like color histograms, texture patterns, edge detection, and shape descriptors. Other computational techniques employed rule-based systems, analyzing specific compositional elements, color palettes, or brushstroke patterns to identify stylistic characteristics. The main focus was to build a reliable algorithm to extract style information from artistic images.
Style attribution and retrieval are related to style transfer, where the goal is to render one image in the style of another.
%Recent research has focused on style transfer between images and style classification. A few studies have addressed in-the-wild style quantification, matching, and retrieval. 
While initially formulated using Markov random fields~\cite{zhang2013style}, style transfer raised in popularity with the rise of deep learning and the work of~\cite{gatys2016image}, which introduced Gram Matrices as style descriptors and an optimization framework for style transfer. Many variants have been proposed, including various regularizations to prevent distortions in reconstructed images~\cite{luan2017deep} or representations of correlation beyond Gram matrices~\cite{chu2018image}. However, recent attribution work shows that Gram-style representations cannot guarantee effective style attribution~\cite{somepalli2024measuring}.

Our work is inspired by more recent advances in style transfer using Generative Adversarial Networks (GANs), where methods like  Adaptive Instance Normalization (AdaIN), which aligns the mean and variance of content features with those of style features, have been shown beneficial for real-time style transfer~\cite{ huang2017arbitrary}. These ideas are central to the StyleGAN~\cite{karras2019style}, where AdaIN layers are embedded into the generator architecture. This allows fine-grained control over different aspects of the generated images, from coarse features like pose to fine details like color schemes.
More recently, diffusion-based style transfer has  been implemented through AdaIN layer manipulation of initial timestep latent~\cite{chung2024style}, and DDIM Inversion-based style latent reuse~\cite{mokady2023null}. These approaches informed our investigation into diffusion model latent distributions.

\subsection{Style-aware Text-to-Image Models}

Early work in style-aware text-to-image generation built upon the success of neural style transfer techniques. \citet{reed2016generative} introduced the first GAN-based text-to-image model, while~\cite{zhang2017stackgan} improved image quality and resolution through a stacked architecture. More recently, diffusion models like DALL-E 2~\cite{ramesh2022hierarchical} and Stable Diffusion~\cite{rombach2022high} have demonstrated impressive capabilities in generating stylized images from text prompts. Several approaches have focused on explicitly representing and manipulating style in text-to-image models, such as style-based GAN architectures with fine-grained style control~\cite{liu2019stgan} or textual inversion methods~\cite{gal2022image, mokady2023null, kumari2023multi} to learn new concepts and styles from a few examples, enabling personalized text-to-image generation. All these works require some fine-tuning of the generative model, whereas we pursue training-free solutions to the style attribution problem.

\subsection{Data Attribution}
While there is extensive literature on image retrieval, most methods focus on semantic similarity, i.e., the retrieval of images of similar content in terms of objects, scenes, etc. Style is a subtle property not well captured by representations, such as DINO~\cite{caron2021emerging} or CLIP~\cite{radford2021learning}, developed for this purpose. Few techniques have been developed specifically for style retrieval or attribution. \citet{lee2021cosmo} employed two separate neural network modules to facilitate style retrieval for image style and content. Recent works~\cite{ruta2021aladin, somepalli2024measuring, wang2023attribution} instead train attribution models using contrastive objectives based on either Behance Artistic Media (BAM) dataset~\cite{ruta2021aladin}, a subset LAION Aesthetic Dataset~\cite{schuhmann2022laion} or synthetic style pairs. \citet{ruta2021aladin} proposes a dual encoder to separate content and style with style features obtained using AdaIN layers, whereas CSD~\cite{somepalli2024measuring} and GDA~\cite{wang2023attribution} fine-tune a pre-trained CLIP or DINO backbone with a linear layer on top with the former training the entire network and latter training the linear layer alone. It is also possible to formulate data attribution as the unlearning of target images~\cite{wang2024data}, but this requires a different model for each target. Unlike these methods, we seek an introspective solution to style attribution that does not require external models or attribution-specific training.
%Compared to the existing data attribution methods, which involve fine-tuning, we build a completely training-free approach that makes our method readily deployable without any additional overheads.

\subsection{Diffusion Features}

In the diffusion model literature, various works have studied the representations learned for denoising, showing that different layers of the network learn features that capture different image properties, like structure, color, or texture~\cite{voynov2023P+, zhang2023prospect, sharma2024alchemist, mukhopadhyay2024text}. Denoising Diffusion Autoencoders (DDAE)~\cite{xiang2023denoising} leverage intermediate activations from pre-trained diffusion models for downstream tasks, demonstrating effectiveness in image classification through linear probing and fine-tuning. Recent advances include REPresentation Alignment (REPA)~\cite{yu2024representation}, which enhances semantic feature learning by aligning diffusion transformer representations with self-supervised visual features. Additionally, diffusion model features have shown promise in zero-shot applications like correspondence~\cite{tang2023emergent}, segmentation~\cite{tian2024diffuse} and classification~\cite{mukhopadhyay2023diffusion}. These observations have inspired our hypothesis that style attribution can be solved introspectively. 

\section{Method}
\label{sec:formatting}
% In this section, we introduce the \ours method for introspective style attribution.

\begin{figure*}[t]
  \centering
   \includegraphics[width=\textwidth]{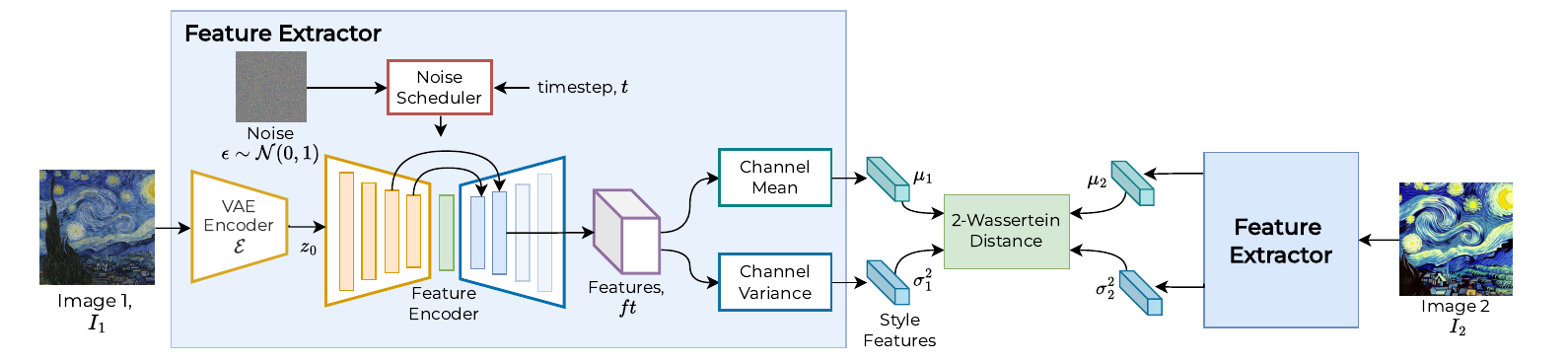}

   \caption{\ours computation of style similarity: channel-wise mean $\mu$ and variance $\sigma^2$ are computed for the identified style layer features. Then 2-Wasserstein Distance is used to measure style similarities between a pair of images.}
   \label{fig:main}
\end{figure*}

\subsection{Diffusion}

Diffusion models~\cite{song2020score} are probabilistic models that define a Markov chain from which images are sampled by gradually denoising a normally distributed seed, using a neural network  $\epsilon_\theta$. T2I models condition this sampling on a text prompt.  A sentence describing the picture is encoded into a text embedding $c$ by a text encoder and then used to condition the denoising network, usually via cross-attention. Stable Diffusion~\cite{rombach2022high}, a prominent T2I model, learns the chain in a latent space of a pre-trained image autoencoder. An encoder $\mathcal{E}$ first maps an image into a latent code $z_0$. In a forward process, a sequence of noisier codes  $z_t$ are obtained by adding Gaussian noise to  $z_0$ according to 
\begin{equation}
z_t = \sqrt{\bar{\alpha_t}}z_0 + \sqrt{1 - \bar{\alpha_t}}\epsilon_t, \quad \epsilon_t \sim N(0,I)
\label{eq:scheduler}
\end{equation}
where $t \in \{0,1,\ldots,T\}$ is a timestep, $T = 1000$, $\bar{\alpha_t} = \prod_{k=1}^t \alpha_k$ and $\alpha_i$ are pre-defined according to a noise schedule.
In a reverse process, the noise latent $z_T$ is used as the seed for a denoising chain, based on the network $\epsilon_\theta$, which aims to recover the noiseless latent $z_0$. This is finally fed to a decoder $\mathcal{D}$ to recover the image. The network parameters $\theta$ are trained to predict the noise $\epsilon_t$ introduced at each step of the forward process, using the loss
\begin{equation}
L_{SD} = \mathbb{E}_{z,\epsilon_t \sim N(0,I),t} \| \epsilon_t - \epsilon_\theta(z_t,t,c) \|_2^2.
\end{equation}
During inference, the reverse process generates a clean latent $\hat{z}_0$, starting from a random seed $z_T \sim N(0,1)$, conditioned on a text embedding $c$. For stable diffusion, the model $\epsilon_\theta$ has a UNet architecture with $4$ downsample, $1$ mid, and $4$ upsample blocks, as shown in Fig. \ref{fig:main}.

% {\color{red} \paragraph{Classifier-Free Guidance} WHY DO WE NEED THIS?
% Classifier-free guidance~\cite{...} is a technique for improving the quality and controllability of generated samples, by jointly training a conditional and an unconditional diffusion model. During training, the conditioning information is randomly dropped at certain time steps, allowing the model to learn both conditional and unconditional distributions. At inference, the conditional and unconditional score estimates are combined with
% \begin{equation}
% \tilde{\epsilon_\theta}(z_t, t, c) = (1 + w) \cdot \epsilon_\theta(z_t, t, c) - w \cdot \epsilon_\theta(z_t, t, \varnothing)
% \end{equation}
% where $w$ is the guidance scale, $\theta(z_t, t, c)$ the conditional score estimate, and $\theta(z_t, t, \varnothing)$ the unconditional score estimate.
% This improves the trade-off between prompt compliance and diversity of the sampled images.
% }

\subsection{Style attribution}

Style attribution aims to identify in a reference dataset ${\cal D} =\{x_1, \ldots, x_n\}$ of images $x_i$, usually but not necessarily the dataset used to train the diffusion model, the subset of images of style closest to that of a query image $q$, which is usually an image synthesized by the model. Like all retrieval operations, this requires two components: 1) a feature representation ${\cal F}: x \rightarrow f$ that maps each image $x$ into a feature tensor $f$ that captures the stylistic elements of $x$, and 2) a similarity metric $d({\cal F}(q), {\cal F}(x))$ that quantifies the style similarity of the feature tensors extracted from query $q$ and reference $x$ images. This allows retrieving the images in $\cal D$ that are most similar to $q$, which can then be inspected, e.g., to determine copyright violations.

\subsection{Introspective style attribution}\label{sec:method}

Style attribution is not a widely studied problem in the literature. Previous approaches~\cite{wang2023attribution, somepalli2024measuring} attempted to train an external feature representation $\cal F$ or adapt existing feature representations, such as DINO~\cite{radford2021learning}, by addition of parameters specific to the extraction of style features. This has two major drawbacks: it requires computationally intensive large-scale training to cover the diffusion model's sampling distribution, and the external feature extractor $\cal F$ introduces substantial memory and computational overhead—often exceeding the diffusion model's resource requirements.

Introspective style attribution aims to solve the problem without external models by using only the features computed by the denoising network $\epsilon_\theta$
of the diffusion model. This is inspired by recent showings that this network produces feature representations that 1)  disentangle image properties such as structure and color~\cite{zhang2023prospect, voynov2023P+} and 2) are sufficiently informative to solve classical vision problems such as classification~\cite{xiang2023denoising}, correspondence~\cite{tang2023emergent}, or segmentation~\cite{tian2024diffuse} with good performance. This suggests that these features should also be sufficient for style attribution. In what follows, we show that this is indeed the case and can be achieved with very simple feature representations and popular similarity metrics, resulting in the simple \ours approach, based on the operations of Fig.~\ref{fig:main}. The features and similarity metric used by \ours are discussed next.

\begin{figure*}[t!]
    \centering
    \scriptsize
    \begin{tabular}{|c|ccccc|}
    \hline &&&&&\\[-6pt]
    \multirow{4}{*}{\bf Originals} & \multicolumn{5}{c|}{{\bf Synthesized}} \\[1pt]
    % \hline
    \cline{2-6} &&&&&\\[-6pt]
     & \multicolumn{5}{c|}{Semantic} \\
     & Claude Monet's & Pablo Picasso's & Gustav Klimt's & Van Gogh's & Leonid Afremov's\\
      & \it{Impression, Sunrise} & \it{Les Demoiselles} & \it{Portrait of Adele} & \it{Starry Night} & \it{Melody Of The Night} \\
     \hline
     % \hline
     %   \includegraphics[width=0.12\textwidth]{sec/figures/artsplit/van.png}  & 
     %   \includegraphics[width=0.12\textwidth]{sec/figures/artsplit/van_van.png} &
     %    \includegraphics[width=0.12\textwidth]{sec/figures/artsplit/van_claude.png} &
     %    \includegraphics[width=0.12\textwidth]{sec/figures/artsplit/van_picasso.png} & 
     %    \includegraphics[width=0.12\textwidth]{sec/figures/artsplit/leo.png} \\
     %    Van Gogh's \it{Starry Night} & \multicolumn{4}{c|}{Style:Van Gogh} \\
     %    \hline
        \hline
        \includegraphics[width=0.12\textwidth]{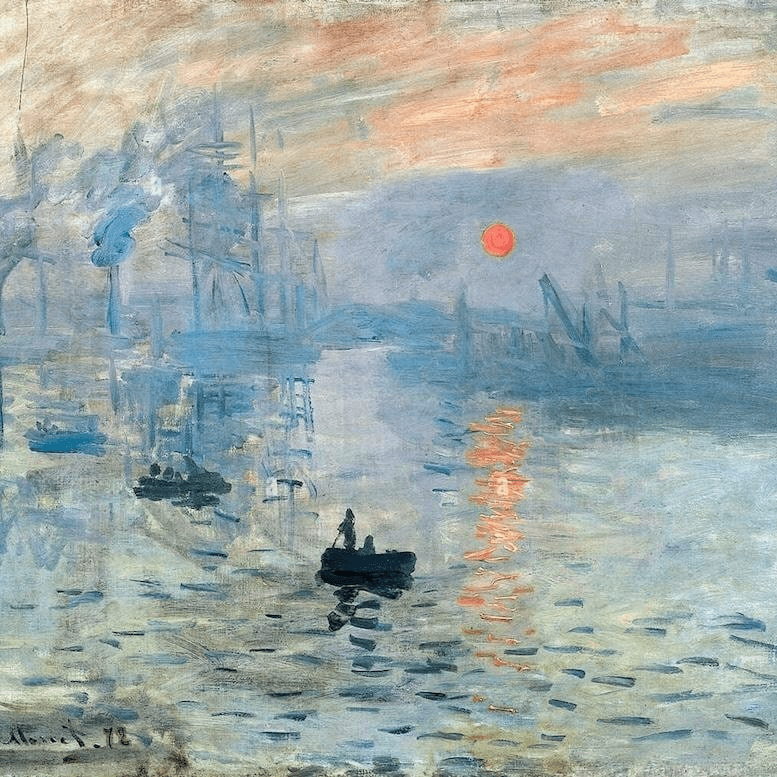} &
        \includegraphics[width=0.12\textwidth]{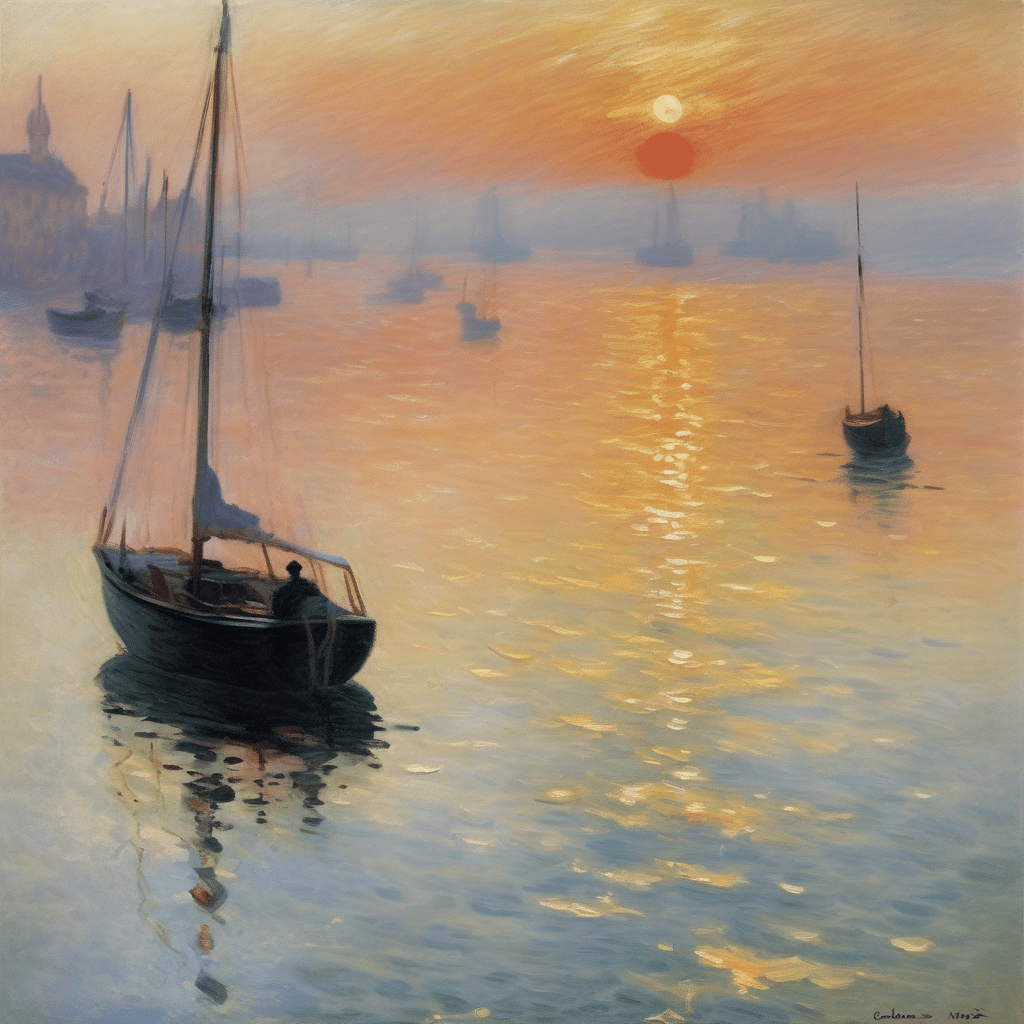} &
        \includegraphics[width=0.12\textwidth]{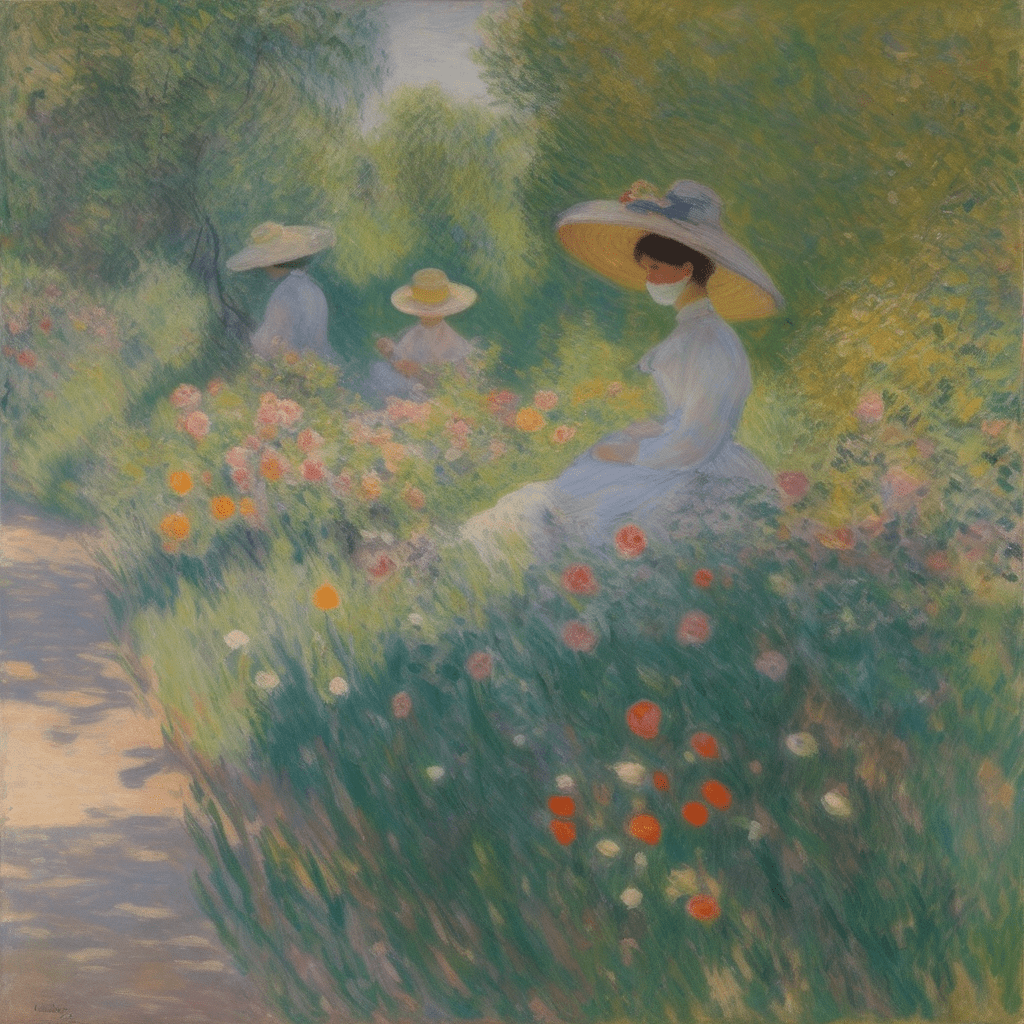} &
        \includegraphics[width=0.12\textwidth]
        {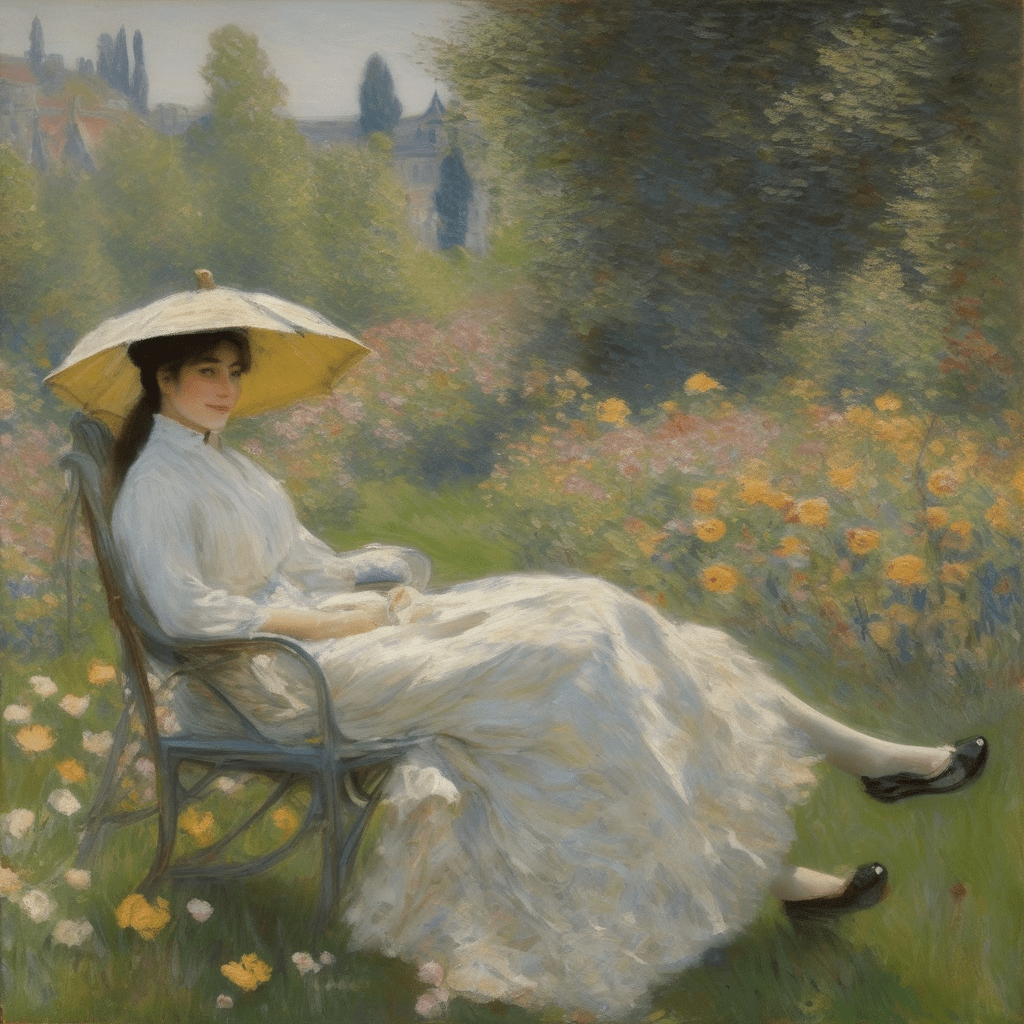} &       \includegraphics[width=0.12\textwidth]{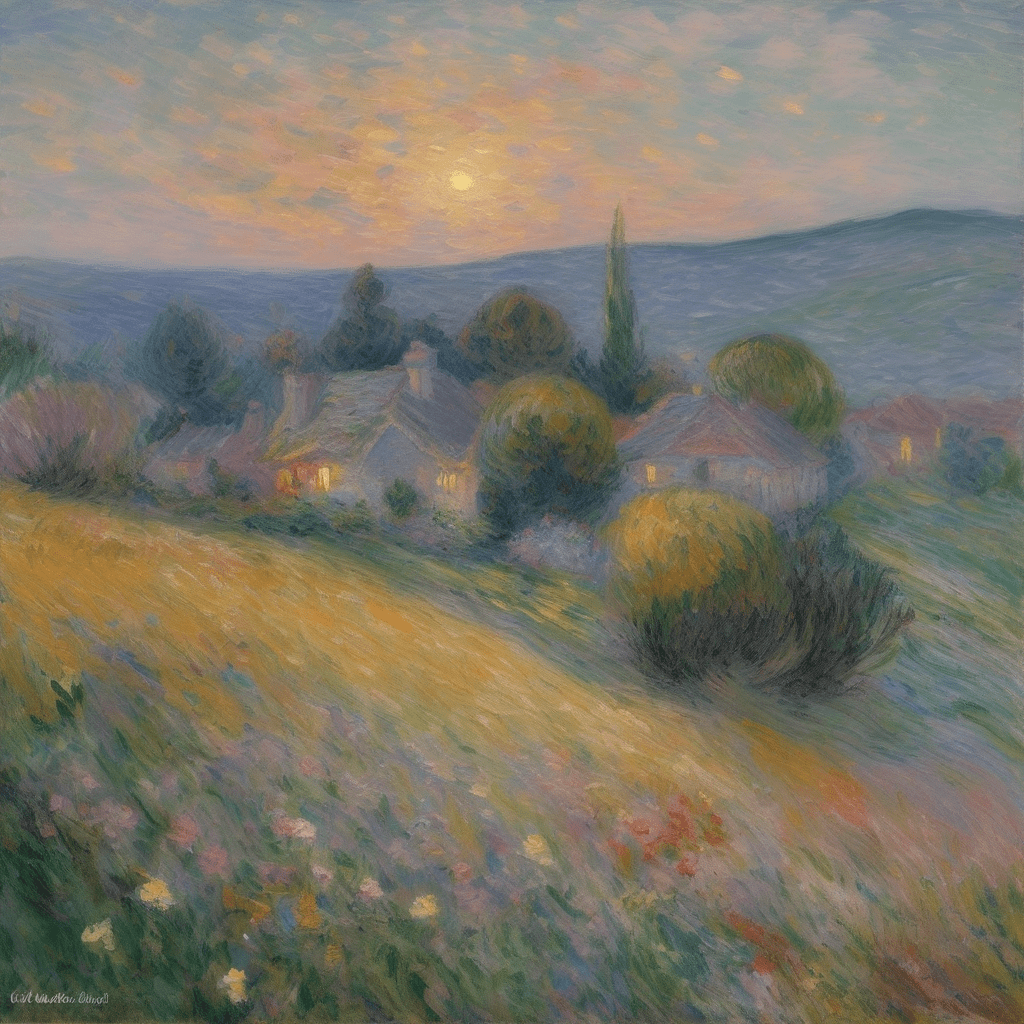} &
        \includegraphics[width=0.12\textwidth]{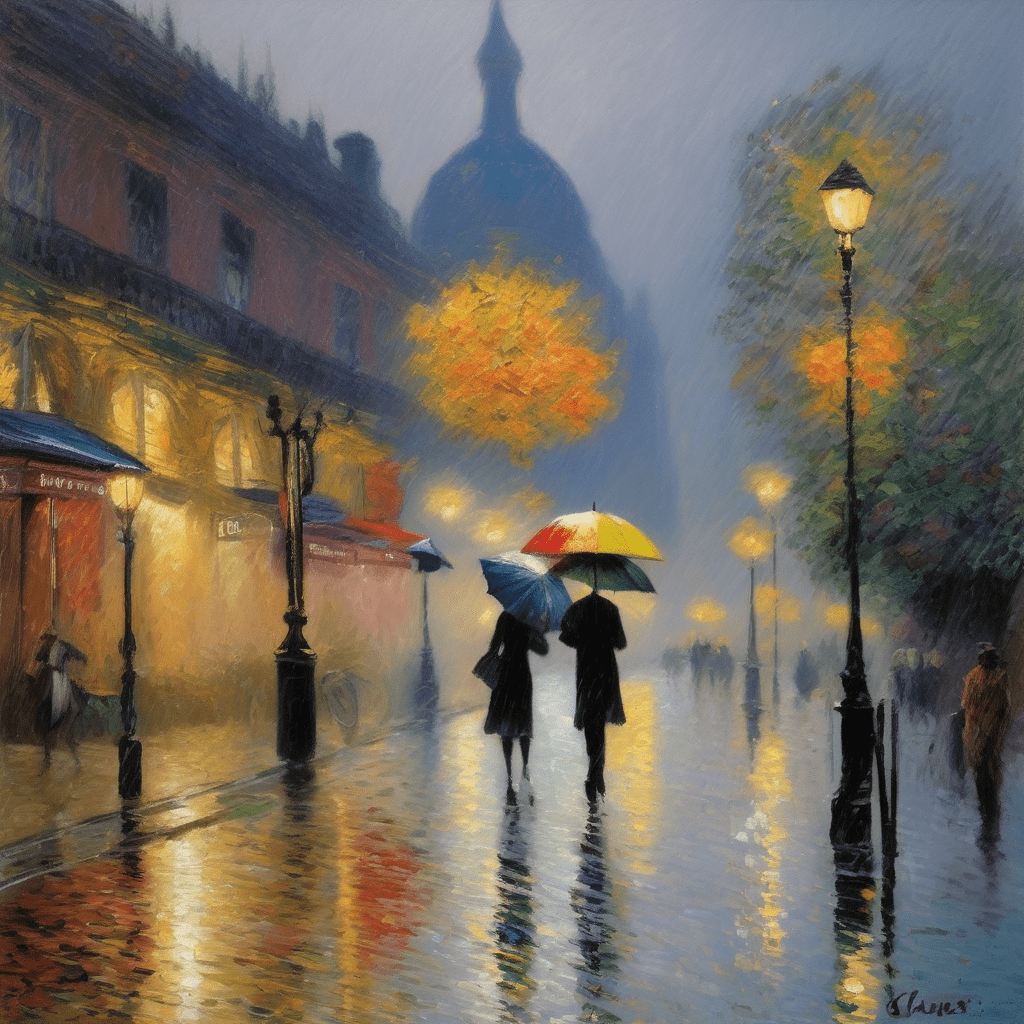} \\
        Claude Monet's \it{Impression, Sunrise} & \multicolumn{5}{c|}{Style:  Claude Monet} \\
        \hline
        \hline
        \includegraphics[width=0.12\textwidth]{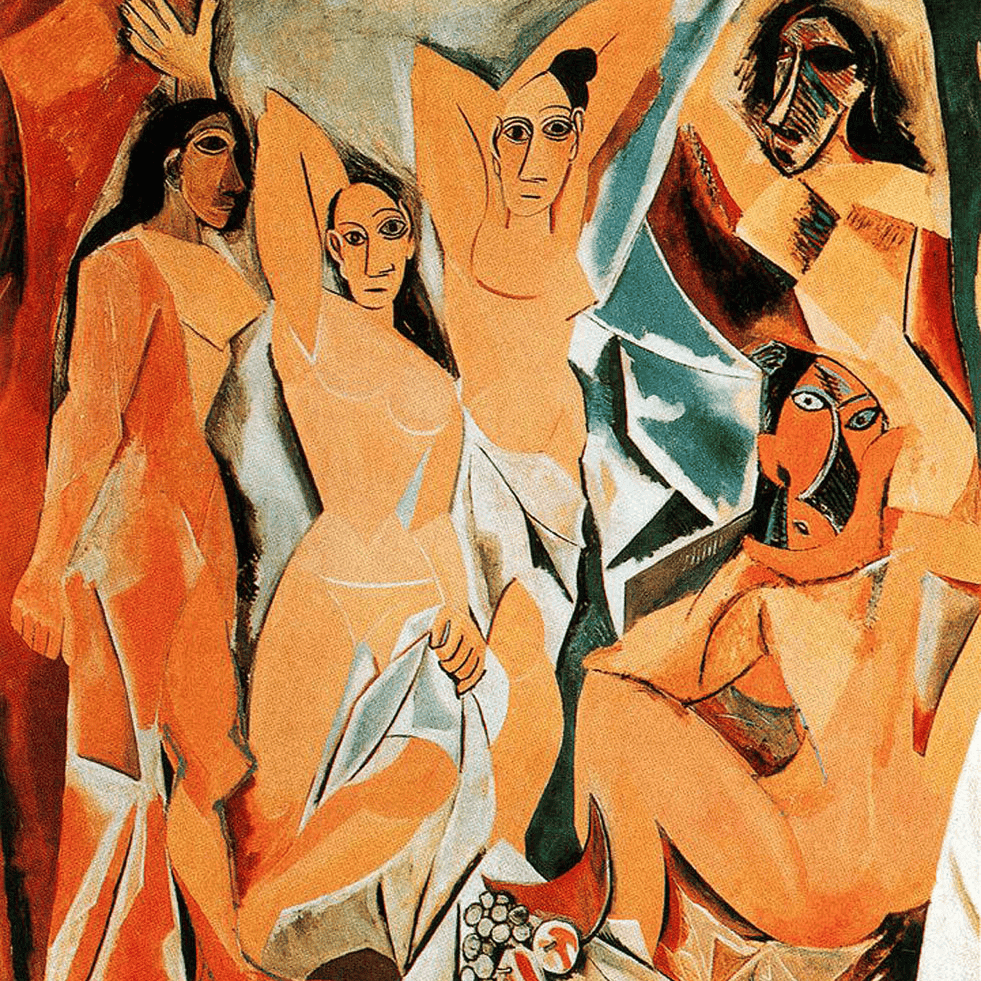} & 
        \includegraphics[width=0.12\textwidth]{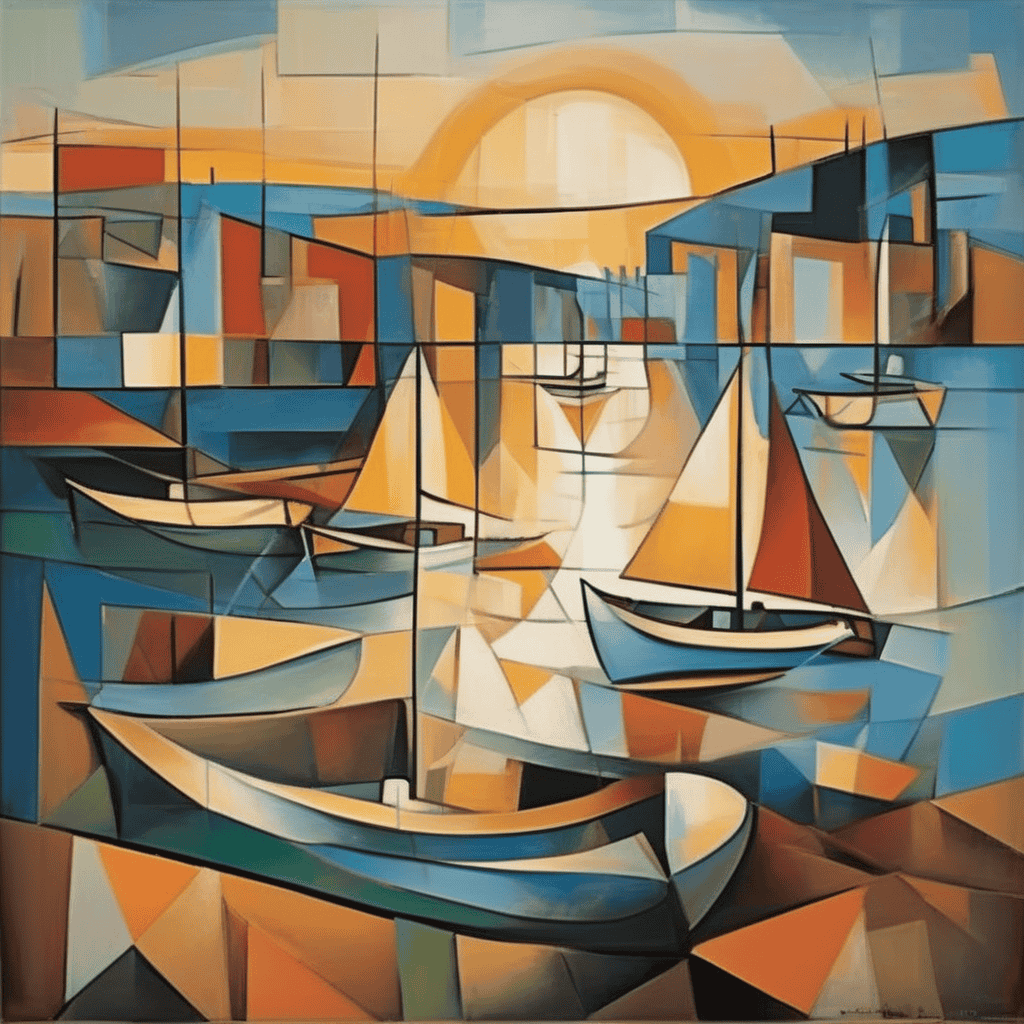} &
        \includegraphics[width=0.12\textwidth]{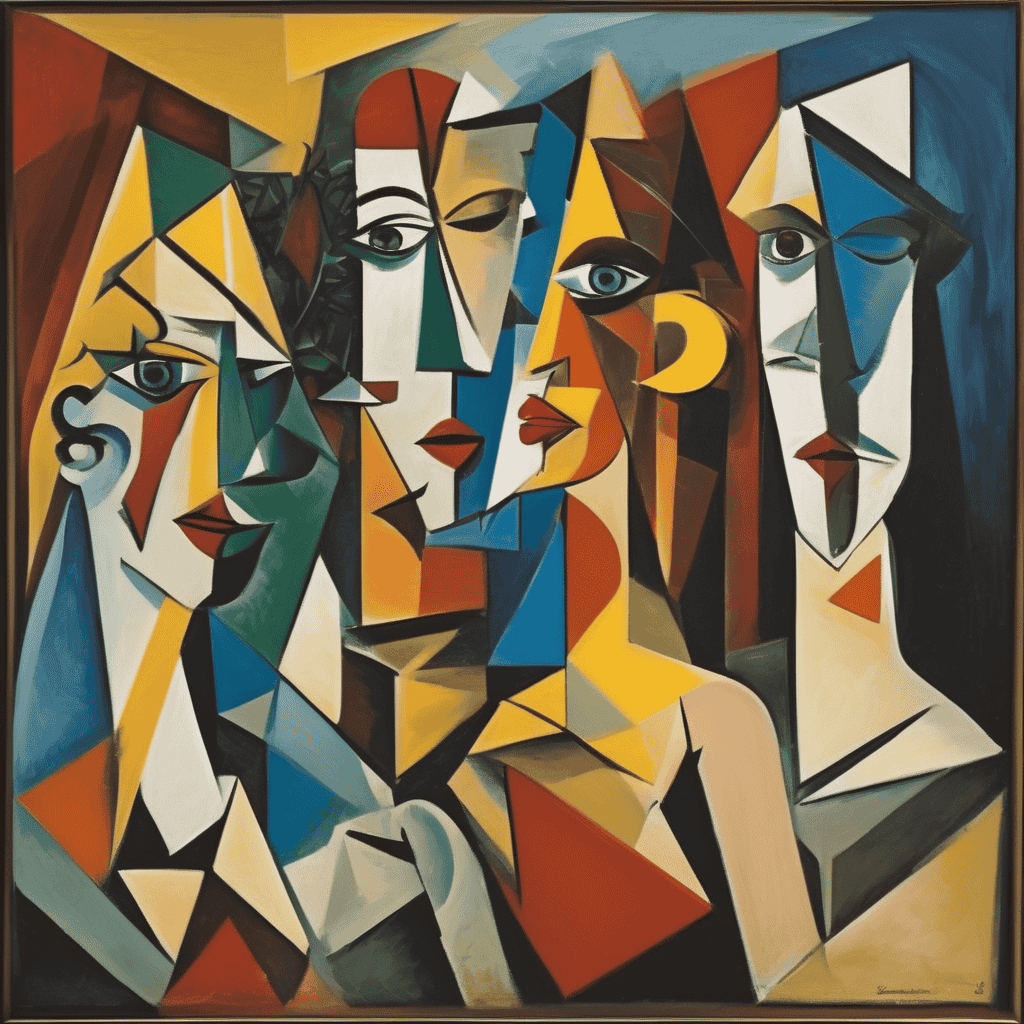} &
        \includegraphics[width=0.12\textwidth]{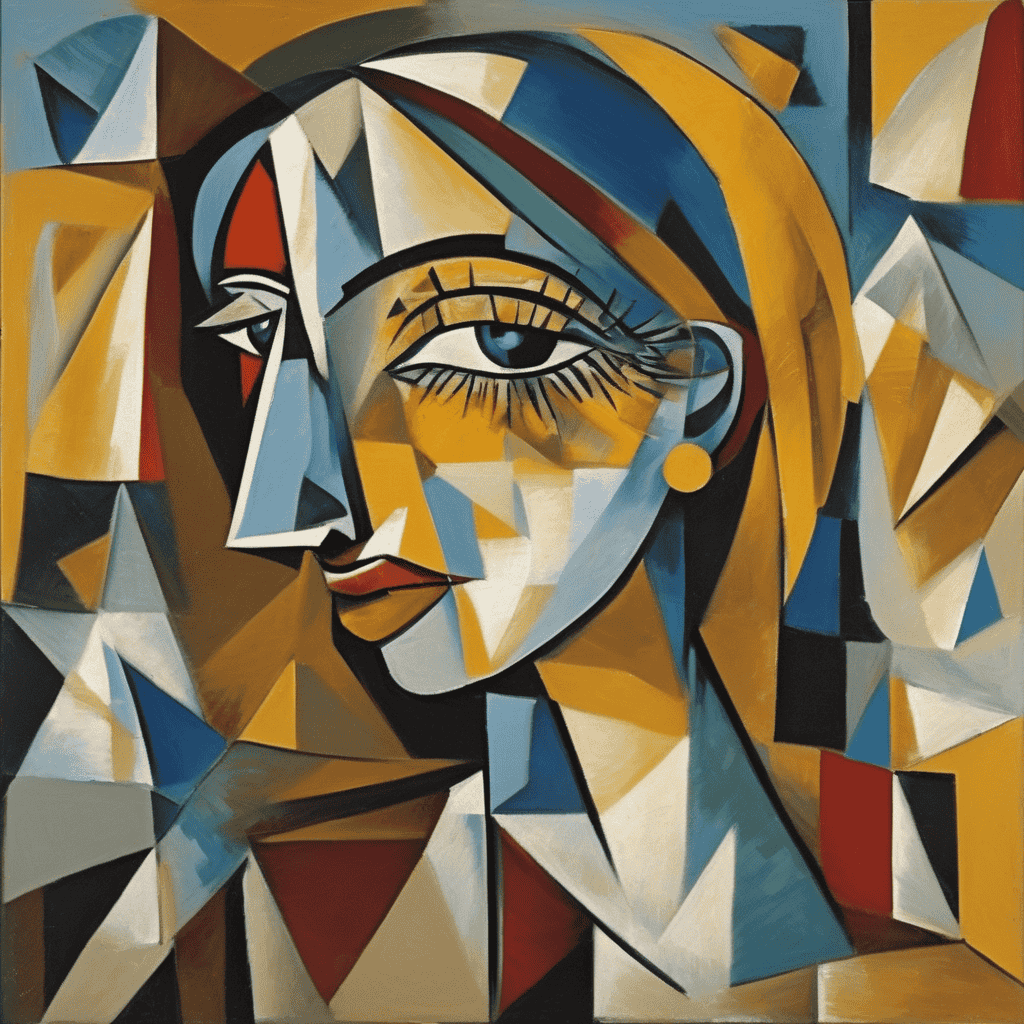} &
        \includegraphics[width=0.12\textwidth]{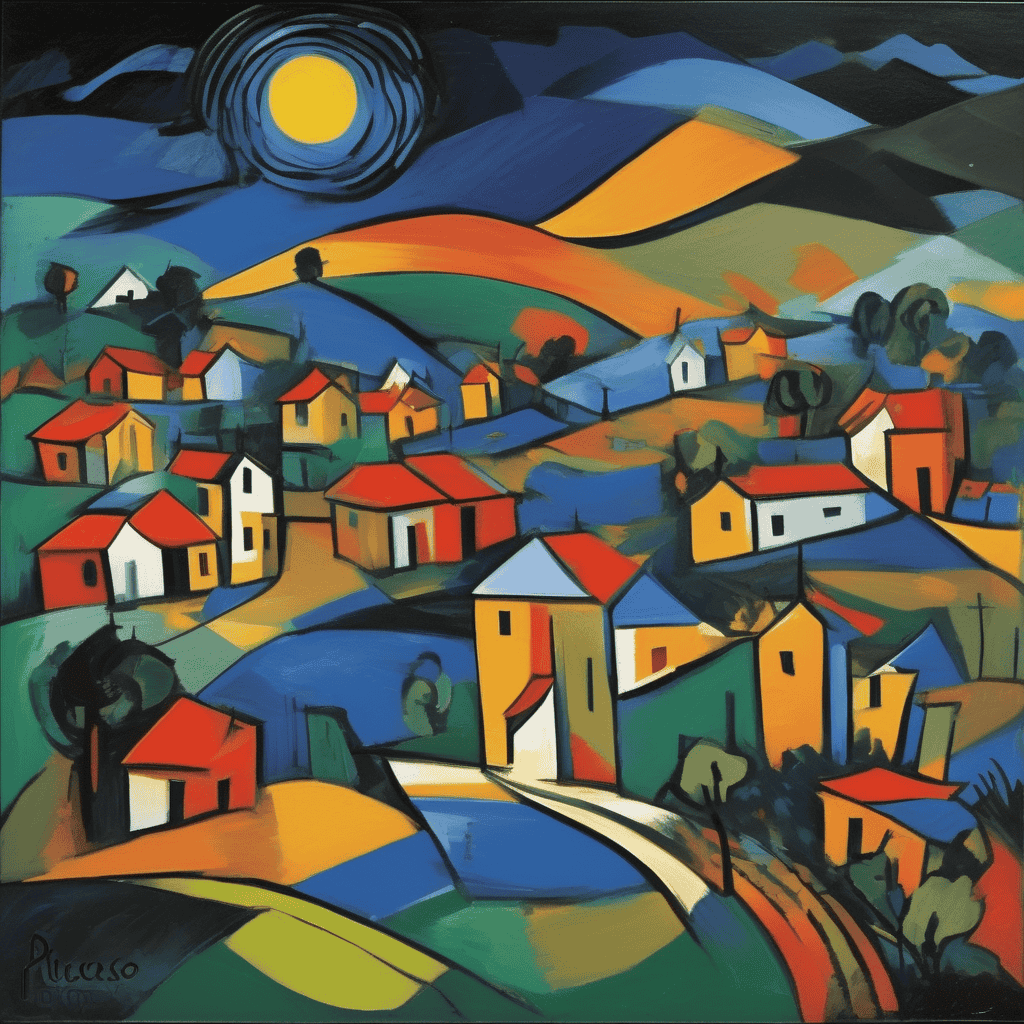} & 
        \includegraphics[width=0.12\textwidth]{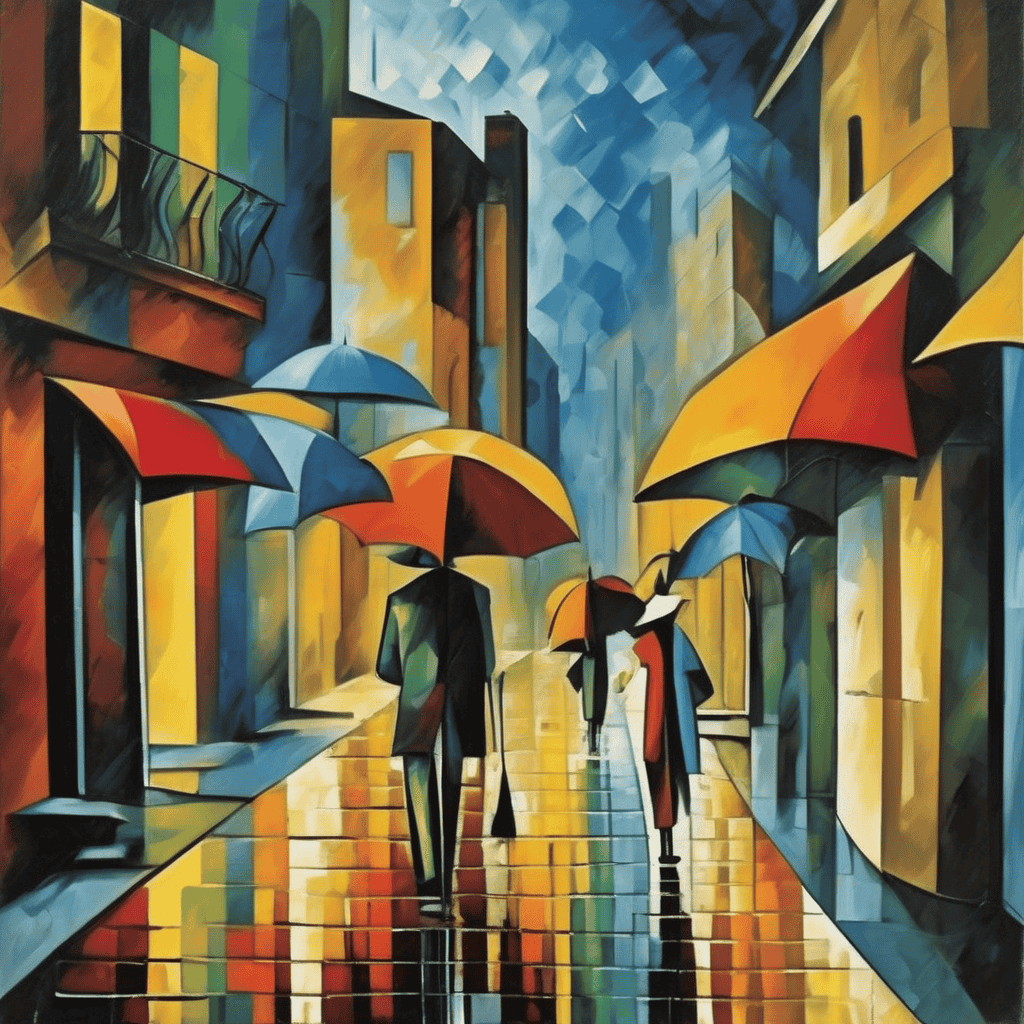}\\
        Pablo Picasso's \it{Les Demoiselles} &  \multicolumn{5}{c|}{Style: Pablo Picasso} \\
        \hline
        \hline
        % \includegraphics[width=0.12\textwidth]{sec/figures/artsplit/leo.png} &
        % \includegraphics[width=0.12\textwidth]{sec/figures/artsplit/leo_van.png} &
        % \includegraphics[width=0.12\textwidth]{sec/figures/artsplit/leo_claude.png} &
        % \includegraphics[width=0.12\textwidth]{sec/figures/artsplit/leo_picasso.png} &
        % \includegraphics[width=0.12\textwidth]{sec/figures/artsplit/leo_leo.png} \\
        % Leonid Afremov's \it{Melody Of The Night} & \multicolumn{4}{c|}{Style:  Leonid Afremov} \\
                \includegraphics[width=0.12\textwidth]{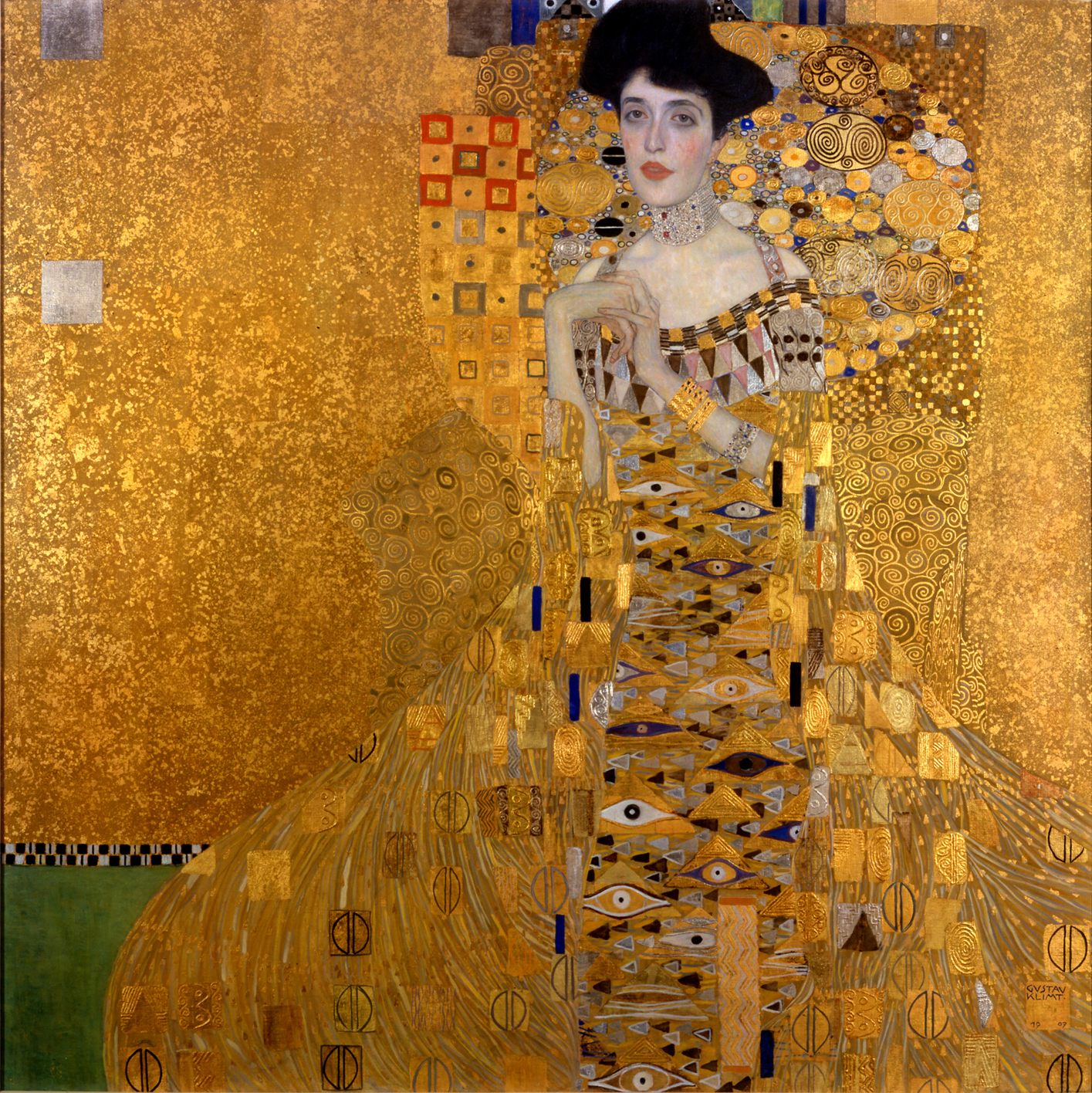} &
        \includegraphics[width=0.12\textwidth]{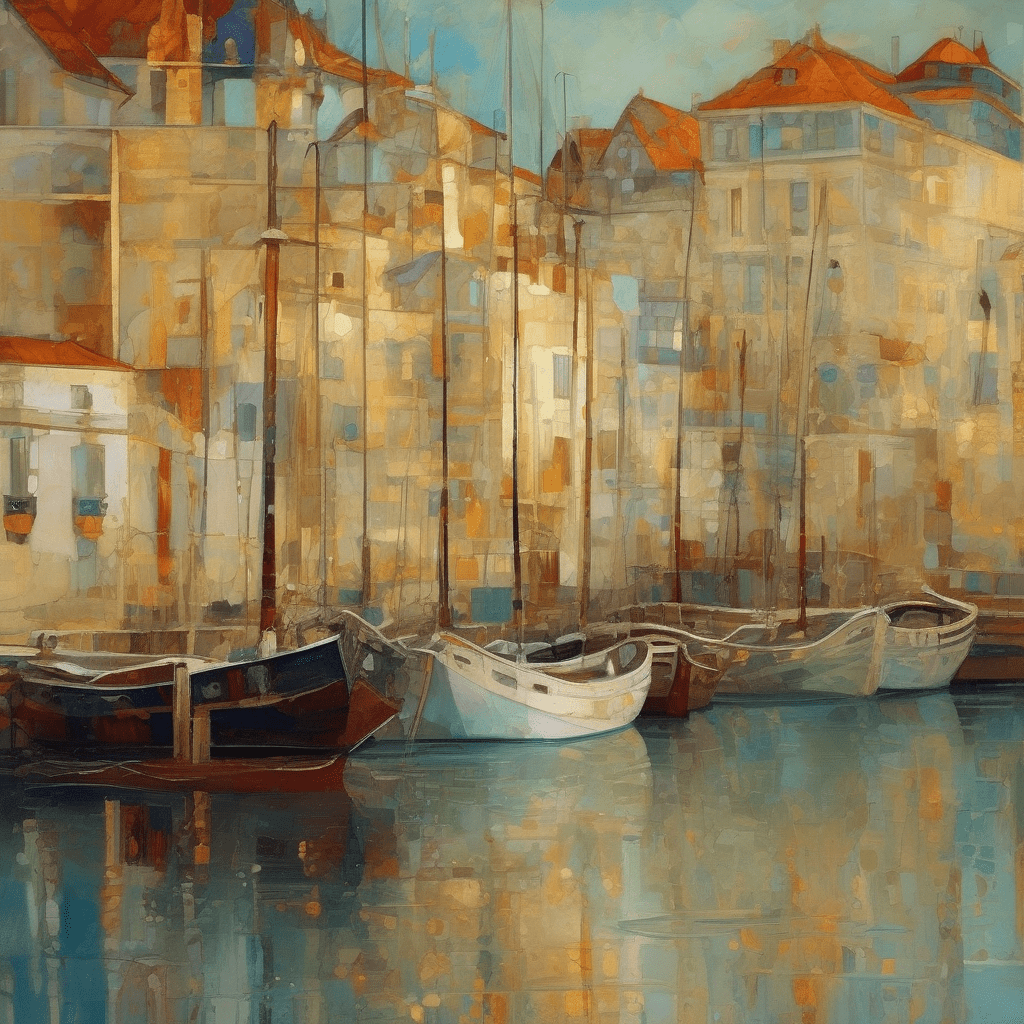} &
        \includegraphics[width=0.12\textwidth]{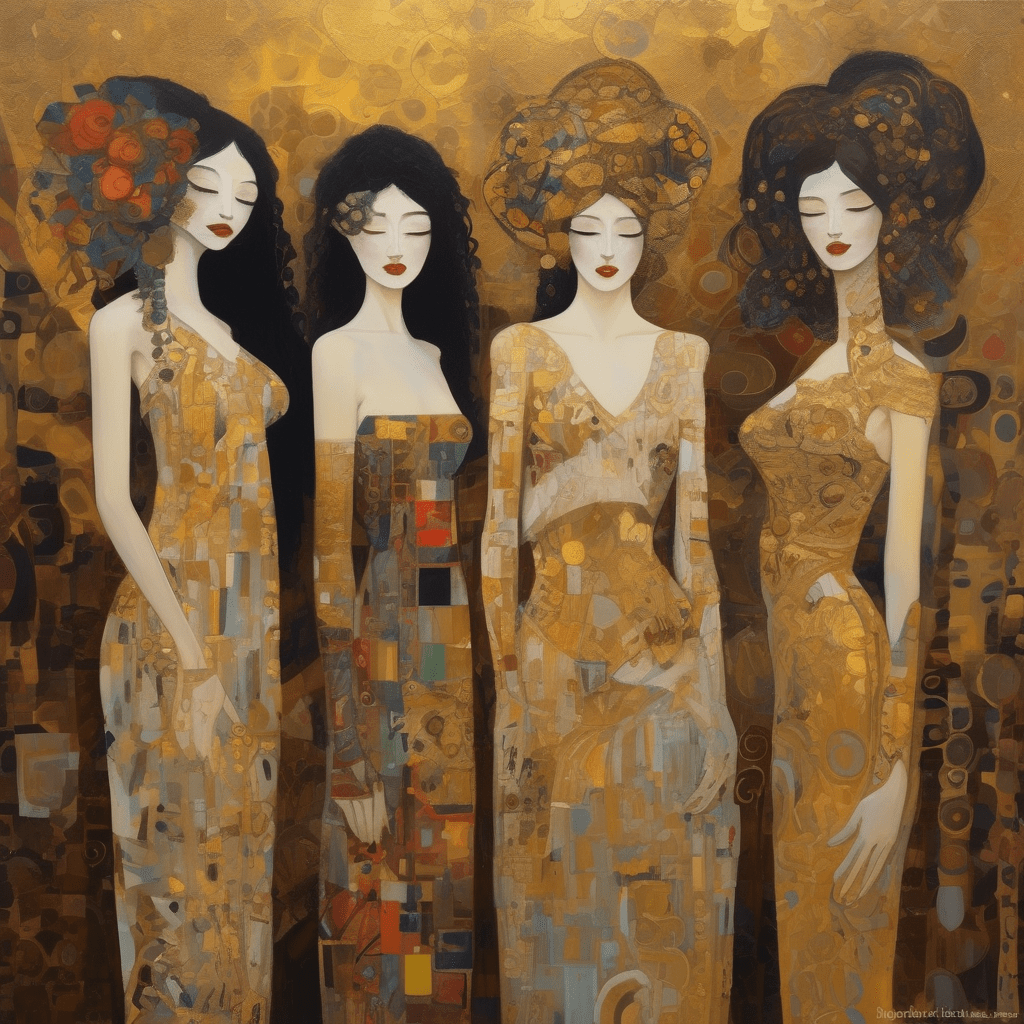} &
        \includegraphics[width=0.12\textwidth]{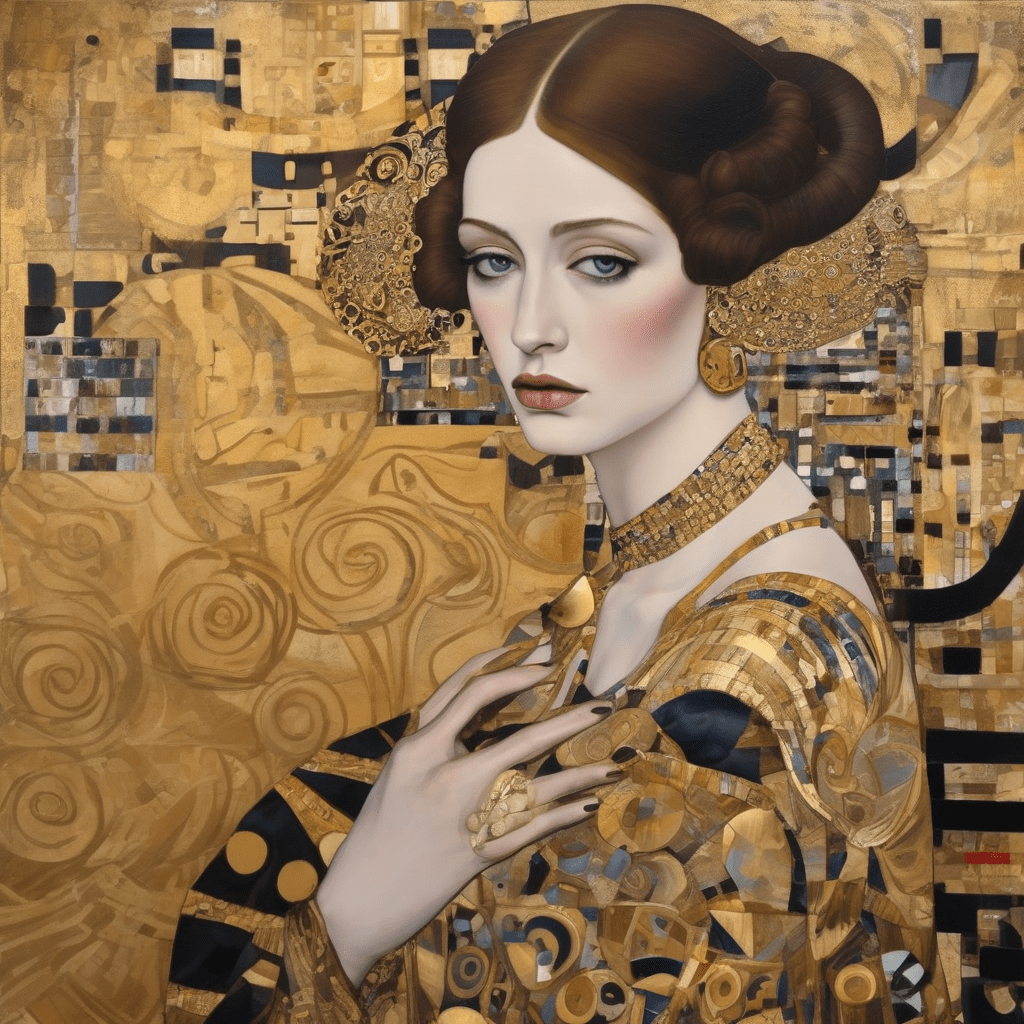}  &
        \includegraphics[width=0.12\textwidth]{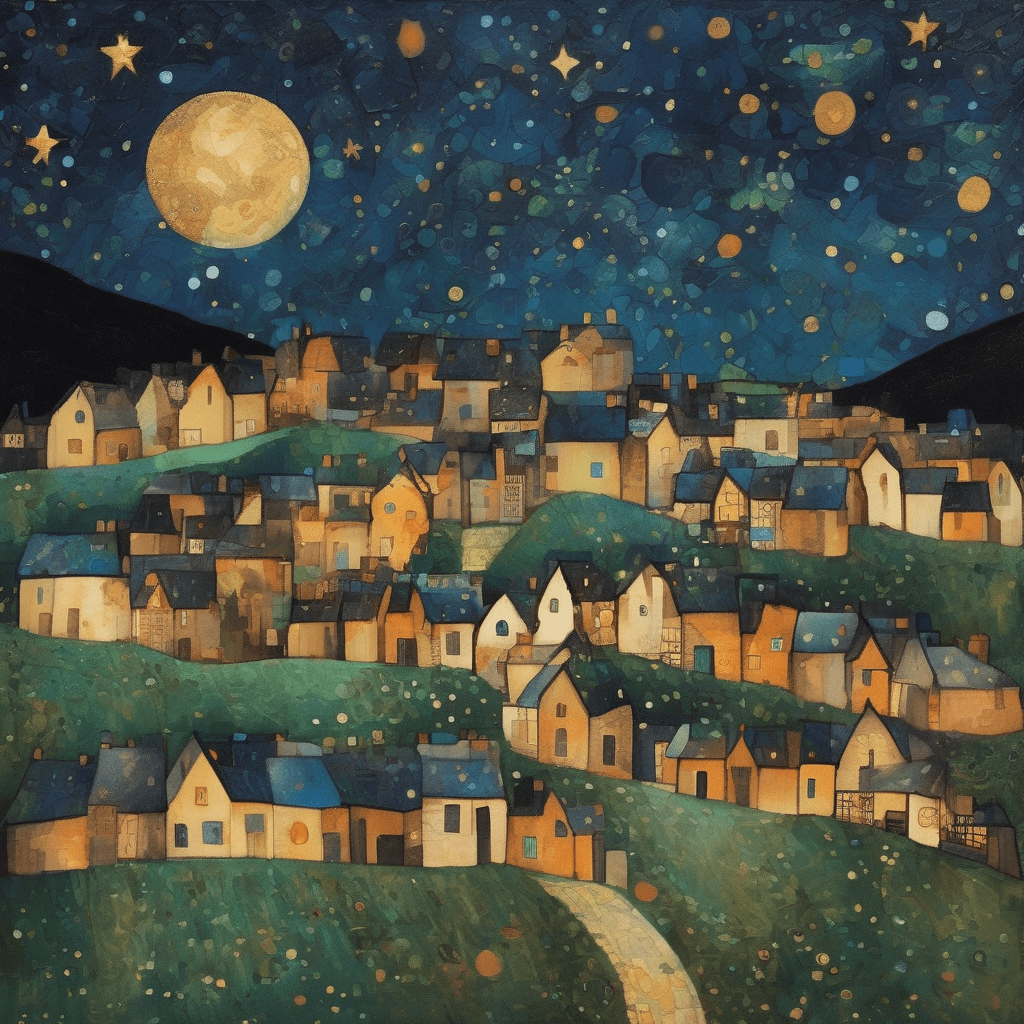} &
        \includegraphics[width=0.12\textwidth]{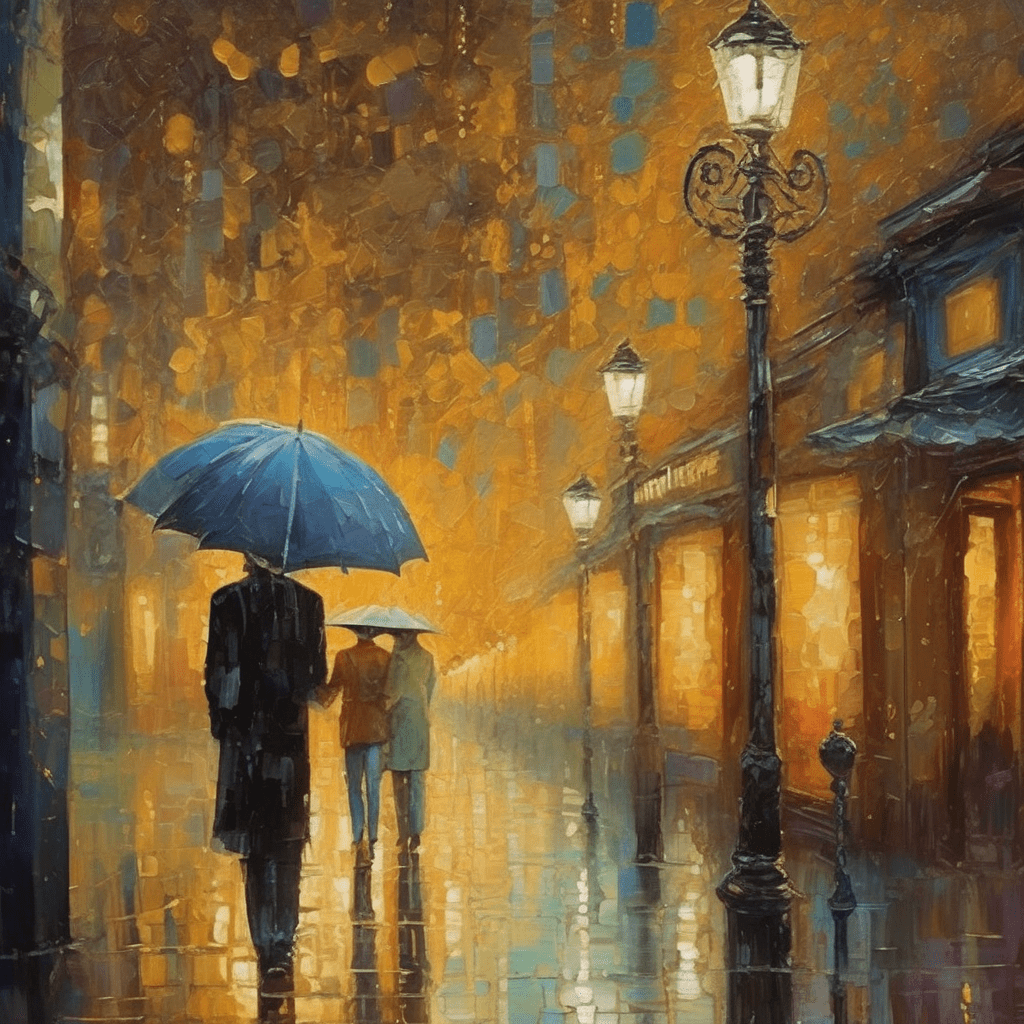}\\
        Gustav Klimt's \it{Portrait of Adele Bloch-Bauer I} & \multicolumn{5}{c|}{Style:  Gustav Klimt} \\
        
        \hline
    \end{tabular}
    \caption{{\it Artistic Style Split} (\ourd) dataset samples.
    Each row shows images generated with the same style, and each column with the same semantics.
    % The images with the same style and semantics as the reference correspond to the diagonals.
    % Each row shows images generated for a reference image by an artist. Reference image, three images generated by ``hacked" prompts, and three images generated by ``innocent" prompts are shown from left to right.
    }
    \label{fig:synth_sample}
\end{figure*}
\def\arraystretch{1}
\paragraph{\ours Features}
To compute a feature representation, image $I$ is first mapped into a latent vector $z_0 = \mathcal{E}(I)$, which is noised to a time $t$ using Eq. \ref{eq:scheduler}. The noised latent $z_t$ is then passed through the denoising network $\epsilon_\theta$ with null text embedding $c = \varnothing$.  A feature tensor $F^t$ is then extracted from intermediate network layers. Given the UNet architecture commonly used to implement $\epsilon_\theta$, we consider this model an autoencoder that stacks a sequence of feature encoder and decoder blocks. The encoder/decoder split is quantified by equating all layers up to the upsampling block $idx$ as an encoder and the remaining upsampling layers as a decoder. The feature tensor $F^{t,idx}$ is finally extracted from block $idx$, as illustrated in Fig.~\ref{fig:main} for $idx = 1$. 

To derive a compact feature representation that captures style information, we then take inspiration from the literature on style-aware generative models and the successful StyleGAN approach to control style by Adaptive Instance Normalization (AdaIN) of first and second-order feature statistics~\cite{huang2017arbitrary}. This leads us to hypothesize that diffusion models may be intrinsically producing similarly normalized features, in which case these statistics should be highly informative of style and eliminate dependence on pixel-wise correlations.
% {\color{red} WASNT THERE ALSO A TALK AT CVPR WHERE THEY DID A SIMILAR NORMALIZATION WITHIN EACH BATCH TO DO SOME STYLE TRANSFER WITH DIFFUSION MODELS?}. 

Based on this intuition, we propose a simple style feature representation for \ours, based on the channel-wise mean and variance of feature responses.
Given the feature tensor $F^{t,idx}$, the mean and variance of channel $c$ are obtained with 
\begin{equation*} 
    \mu_c = \frac{1}{W H} \sum_{i=1}^W \sum_{j=1}^{H} F^{t,idx}_{c,i,j}, \;
    \sigma_c^2 =\frac{1}{W H} \sum_{i=1}^W \sum_{j=1}^{H} (F^{t,idx}_{c,i,j} - \mu_c)^2
\label{eq:mu_sig}
\end{equation*}
where $W,H$ are the spatial dimensions of the tensor. The style feature representation of the image $I$ is finally 
\begin{equation}
    f^{t,idx}(I) = (\mu_1, \ldots, \mu_C, \sigma_1^2, \ldots, \sigma_C^2)^T
\end{equation}
where $C$ is the number of channels at block $idx$. Both $t$ and $idx$ are hyperparameters of the method, which is summarized in Fig.~\ref{fig:main}.

\paragraph{Similarity Metric}
The \ours representation of an image models its feature statistics with a multivariate Gaussian distribution of $C$ dimensions. As shown in Fig.~\ref{fig:main}, Gaussian parameters $(\mu_1, \Sigma_1)$ and $(\mu_2, \Sigma_2)$ are extracted for images $I_1$ and $I_2$, respectively. Hence, any similarity measure between two multivariate Gaussians can be used to compare the images. We have investigated several similarity measures
% The $L_2$, or Euclidean, distance 
% \begin{equation}
%     L_2(\mu_1, \mu_2)^2 = \|\mu_1 - \mu_2\|_2^2,
% \end{equation}
% ignores covariance information and does not have an interpretation as a measure of similarity between probability distributions. This is also the case for the Frobenius norm between the Gram matrices  extracted from the two images, which are formed by taking the outer product of their mean vectors $\mu$, 
% \begin{equation}
%     Gram(\mu_1, \mu_2) = \|\mu_1 \mu_1^T - \mu_2\mu_2^T\|_F.
% \end{equation}
% This measure is popular in the style transfer literature. A popular measure of similarity between distributions is the  Kullback–Leibler (KL) divergence. For two multivariate Gaussians this has the form
% \begin{eqnarray}
% \lefteqn{\mathrm{KL}((\mu_1, \Sigma_1) || (\mu_2, \Sigma_2)) =} &&\\
% && \frac{1}{2} \left[ \log \frac{|\Sigma_2|}{|\Sigma_1|} + \mathrm{tr}(\Sigma_2^{-1}\Sigma_1) + (\mu_2 - \mu_1)^T \Sigma_2^{-1} (\mu_2 - \mu_1) \right] \nonumber.
% \end{eqnarray}
% However, the KL divergence is not symmetric. The Jensen-Shannon Divergence 
% \begin{equation}
%     \mathrm{JSD}(I_1 || I_2) = \frac{1}{2}\mathrm{KL}(I_1 || I_2) + \frac{1}{2}\mathrm{KL}(I_2 || I_1)
% \end{equation}
% eliminates this problem.
and found the 2-Wasserstein ($W_2$) distance 
\begin{eqnarray}
    \lefteqn{W_2((\mu_1, \Sigma_1); (\mu_2, \Sigma_2))^2  = \|\mu_1 - \mu_2\|_2^2 } && \\ 
    & & + \,\, \text{tr} \left(\Sigma_1 + \Sigma_2 - 2(\Sigma_1^{1/2}\Sigma_2\Sigma_1^{1/2})^{1/2}\right). \nonumber
\end{eqnarray}
to be effective for style retrieval. A comparison to other metrics, including Euclidean ($L_2$) Distance, Gram Matrix, and Jensen-Shannon Divergence ($JSD$), is presented in the experiments. In all cases, the covariance matrices are assumed diagonal with the variances of the \ours representation as diagonal entries.

\begin{figure*}[t!]
    \centering
    \begin{subfigure}[b]{0.1\textwidth}
        \centering
        \includegraphics[width=0.97\textwidth]{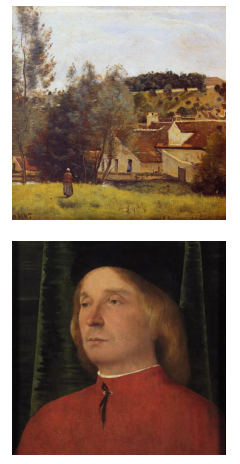}
        \caption{Queries}
        
    \end{subfigure}
    \hfill
    \begin{subfigure}[b]{0.29\textwidth}
        \centering
        \includegraphics[width=1\textwidth]{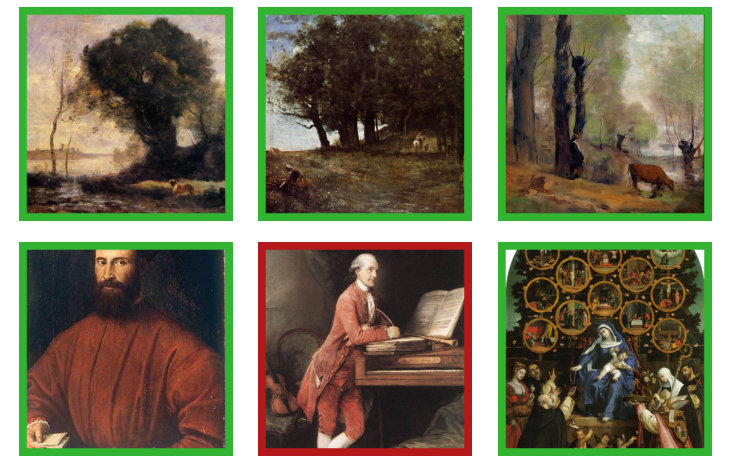}
        \caption{\ours (ours)}
        
    \end{subfigure}
    \hfill
    \begin{subfigure}[b]{0.29\textwidth}
        \centering
        \includegraphics[width=1\textwidth]{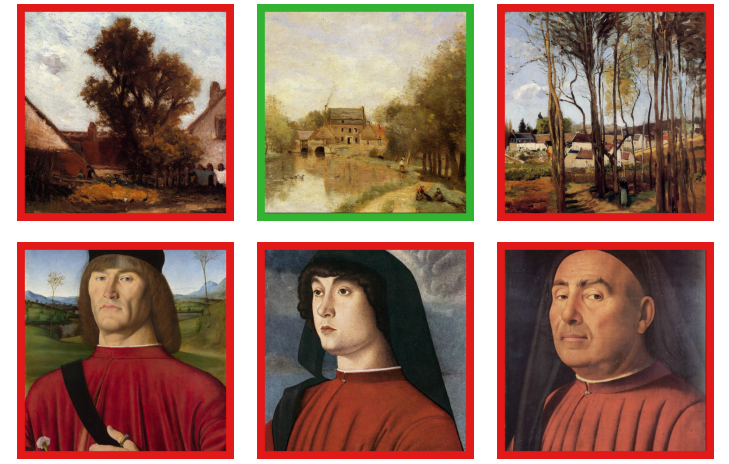}
        \caption{CSD \cite{somepalli2024measuring}}
        
    \end{subfigure}
    \hfill
    \begin{subfigure}[b]{0.29\textwidth}
        \centering
        \includegraphics[width=1\textwidth]{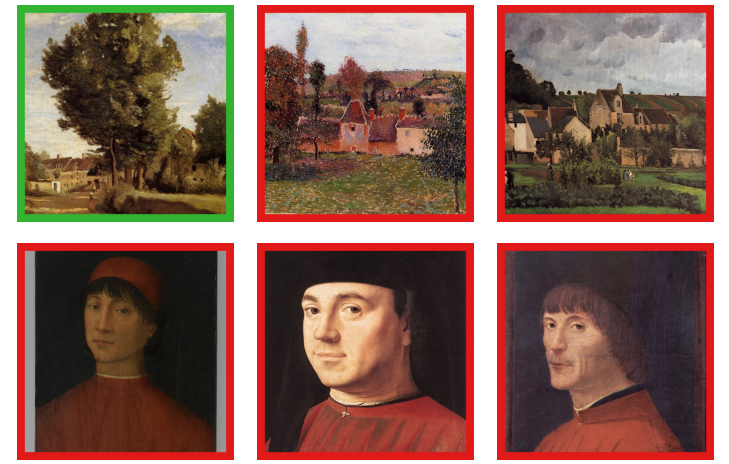}
        \caption{GDA \cite{wang2023attribution}}
        
    \end{subfigure}
    \caption{Image retrieval on WikiArt Dataset for \ours, CSD, and GDA. The first column (left to right) shows the query image, followed by \ours's top 3 retrieval results, CSD and GDA, respectively. Green colors indicate \textcolor{correct}{correct} and red for \textcolor{wrong}{incorrect} retrievals.}
    \label{fig:wikiart_res}
\end{figure*}

\begin{table*}
  \centering
  \footnotesize
  \begin{tabular}{l | c c c | c c c | c c c | c c c}
    \toprule
     & \multicolumn{6} {c|} {WikiArt} & \multicolumn{6}{c}{DomainNet} \\
    \cline{2-13} 
    & \multicolumn{3}{c|}{mAP@$k$} & \multicolumn{3}{c|}{Recall@$k$} & \multicolumn{3}{c|}{mAP@$k$} & \multicolumn{3}{c}{Recall@$k$}\\
    Method & 1 & 10 & 100 & 1 & 10 & 100 & 1 & 10 & 100 & 1 & 10 & 100 \\
    \midrule
    VGG-Net Gram \cite{gatys2016image} & 0.259 & 0.194 & 0.114 & 0.259 & 0.527 & 0.804 & - & - & - & - & - & - \\
    SSCD ResNet-50 \cite{pizzi2022self} & 0.332 & 0.243 & 0.135 & 0.332 & 0.591 & 0.839 & 0.713 & 0.672 & 0.608 & 0.713 & 0.964 & \cellcolor{tabfirst} 0.998 \\  
    MoCo ViT-B/16 \cite{he2019momentum} & 0.440 & 0.332 & 0.188 & 0.440 & 0.689 & 0.879 & 0.784 & 0.757 & 0.708 & 0.784 & 0.960 &\cellcolor{tabsecond} 0.997 \\ 
    DINO ViT-B/16 \cite{caron2021emerging} & 0.458 & 0.348 & 0.197 & 0.458 & 0.706 & 0.889 & 0.774 & 0.741 & 0.684 & 0.774 & 0.956 & \cellcolor{tabsecond}0.997 \\ 
    CLIP ViT-L \cite{radford2021learning}& 0.595 & 0.489 & 0.316 & 0.595 & 0.830 & \cellcolor{tabsecond} 0.952 & \cellcolor{tabsecond} 0.928 &  \cellcolor{tabsecond} 0.923 & \cellcolor{tabsecond} 0.778 & \cellcolor{tabsecond} 0.928 & \cellcolor{tabsecond}0.972 & 0.996 \\
    \midrule
    GDA ViT-B \cite{wang2023attribution} & 0.439 & 0.331 & 0.186 & 0.439 & 0.685 & 0.873 & 0.800 & 0.768 & 0.694 & 0.800 & 0.966 & \cellcolor{tabfirst} 0.998 \\ 
    GDA CLIP ViT-B \cite{wang2023attribution} & 0.522 & 0.419 & 0.259 & 0.522 & 0.782 & 0.935 & 0.817 &  0.786 & 0.716 & 0.817 &  \cellcolor{tabsecond}0.972 & \cellcolor{tabfirst} 0.998\\
    GDA DINO ViT-B \cite{wang2023attribution} & 0.458 & 0.349 & 0.199 & 0.458 & 0.709 & 0.8888 & 0.770 & 0.739 & 0.684 & 0.770 & 0.956 & \cellcolor{tabsecond}0.997\\
    CSD ViT-L \cite{somepalli2024measuring} & \cellcolor{tabsecond}0.646 & \cellcolor{tabsecond} 0.538 &  \cellcolor{tabsecond}0.357 &  \cellcolor{tabsecond}0.646 & \cellcolor{tabsecond}0.857 & {\cellcolor{tabfirst} 0.956} &  0.833 & 0.811 & 0.766 & 0.833 & \cellcolor{tabsecond}0.972 & {\cellcolor{tabfirst} 0.998} \\
    \midrule
    % StyFeat-Idx: $0$, $t$: 25 & 0.866 & 0.830 & 0.474 & 0.866 & 0.879 & 0.911  & 0.928 & 0.923 & 0.778 & 0.928 & 0.972 & 0.996 \\
    \ours (Ours)  & \cellcolor{tabfirst}0.887 & {\cellcolor{tabfirst} 0.852} & {\cellcolor{tabfirst} 0.525} & {\cellcolor{tabfirst} 0.887} &{\cellcolor{tabfirst} 0.909} & 0.941 & {\cellcolor{tabfirst} 0.954} & {\cellcolor{tabfirst} 0.949} & {\cellcolor{tabfirst} 0.850} & {\cellcolor{tabfirst} 0.954} & {\cellcolor{tabfirst} 0.982} & 0.995 \\ 
    % StyFeat-Idx: $2$, $t$: 25 & 0.888 & 0.852 & 0.521& 0.888 & 0.907 & 0.938 & 0.952 & 0.948 & 0.847 & 0.952 & 0.981 & 0.996  \\ 
    % StyFeat-Idx: $3$, $t$: 25 & 0.872 & 0.836 & 0.491& 0.872 & 0.886 & 0.916  & 0.935 & 0.934 & 0.812 & 0.935 & 0.974 & 0.994 \\ 
    \bottomrule
  \end{tabular}
  \caption{Evaluation on WikiArt and DomainNet datasets. \ours outperforms all previous methods significantly. The best values are highlighted with \colorbox{tabfirst}{red} and second best with \colorbox{tabsecond}{orange}.}
  \label{tab:main}
\end{table*}

\section{The \ourd Dataset}

% \textcolor{red}{the logic can follows the previous version, but the contexts can be adjusted -- 1) why we need the sHacks Dataset; 2) how you create the dataset?, maybe also refer to fig.??? }

To address the limitations of existing datasets for fine-grained evaluation of style retrieval, we propose the {\it Artistic Style Split} (\ourd) dataset. This was created with the prompt-image pairs of the $2$ most recognized works of $50$ prominent artists from the LAION Aesthetic Dataset \cite{schuhmann2022laion}. For each of the two paintings, ChatGPT\cite{openai2023chatgpt} was asked to generate a ``style" specification and a ``semantic" description, such that there is no style information in the semantic description and vice-versa. Stable Diffusion v2.1 was then used with a combination of two prompts, ``style" and ``semantic," to synthesize a reference image dataset. With 50 artists (``styles") and 100 paintings (``semantics"), this led to $50 \times 100 = 5,000$ prompt combinations. A set of $12$ images was sampled per combination, yielding $60,000$ images in total. The procedure is detailed in Supplemental Section D. 

Fig.~\ref{fig:synth_sample} shows some of the images in the dataset. The left column shows three of the $100$ original paintings used to produce the style and semantic descriptions, e.g. style: ``Gustav Klimt" and semantic: ``Portrait of Adele''. The remaining table entries are images synthesized for different combinations of style (row) and semantic (column). Note how the procedure creates a fine-grained sample of the style $\times$ semantic space. 

Style attribution performance is measured using the original artist's paintings (left column at Fig.~\ref{fig:synth_sample}) to retrieve images. Each original is used as a query, the dataset images are ranked by similarity to this query, and performance is evaluated by the agreement between the labels of the retrieved images and that of the query. This is measured using a weighted score of Top-k retrieved neighbors that are correctly matched. Based on the labels, we perform two assessments on the retrieved images. Style labels allow the evaluation of style attribution accuracy. On the other hand, semantic labels allow the evaluation of semantic similarity and reveal the model's semantic bias. A good representation for style attribution should simultaneously achieve high style and low semantic retrieval accuracy. 

\begin{figure*}[t!]
    \centering
    \begin{subfigure}[b]{0.1\textwidth}
        \centering
        \includegraphics[width=0.97\textwidth]{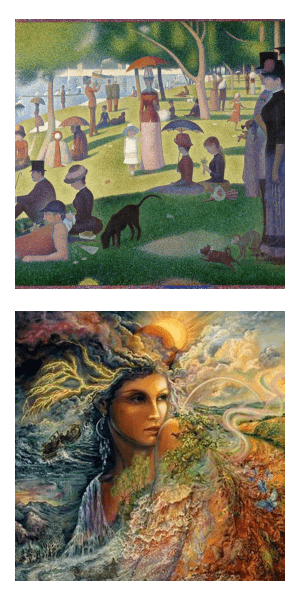}
        \caption{Queries}
        
    \end{subfigure}
    \hfill
    \begin{subfigure}[b]{0.29\textwidth}
        \centering
        \includegraphics[width=1\textwidth]{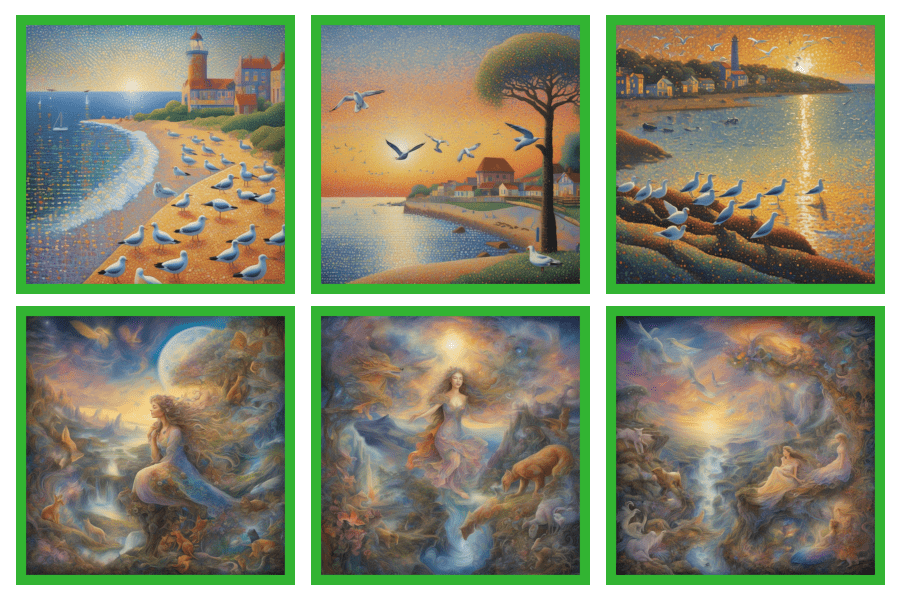}
        \caption{\ours (ours)}
        
    \end{subfigure}
    \hfill
    \begin{subfigure}[b]{0.29\textwidth}
        \centering
        \includegraphics[width=1\textwidth]{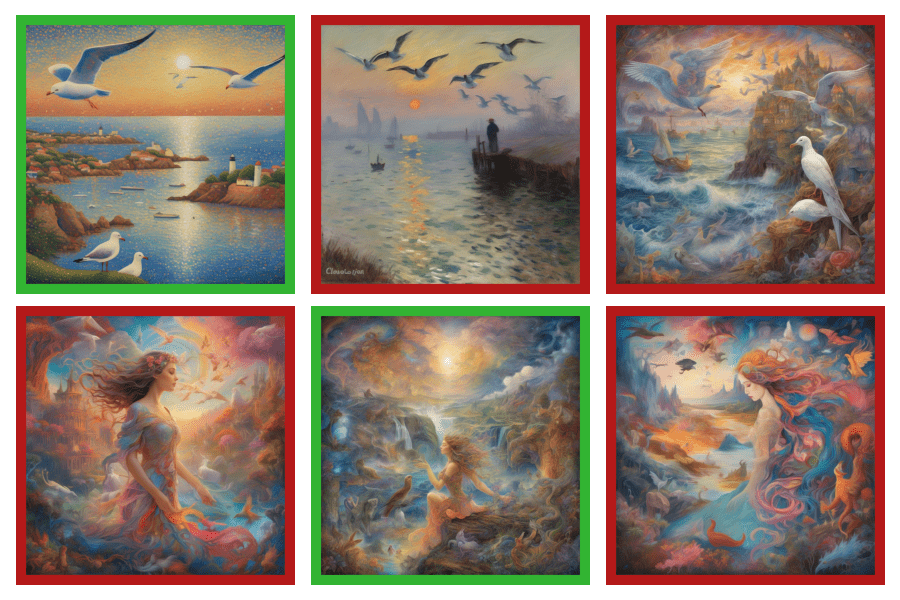}
        \caption{CSD \cite{somepalli2024measuring}}
        
    \end{subfigure}
    \hfill
    \begin{subfigure}[b]{0.29\textwidth}
        \centering
        \includegraphics[width=1\textwidth]{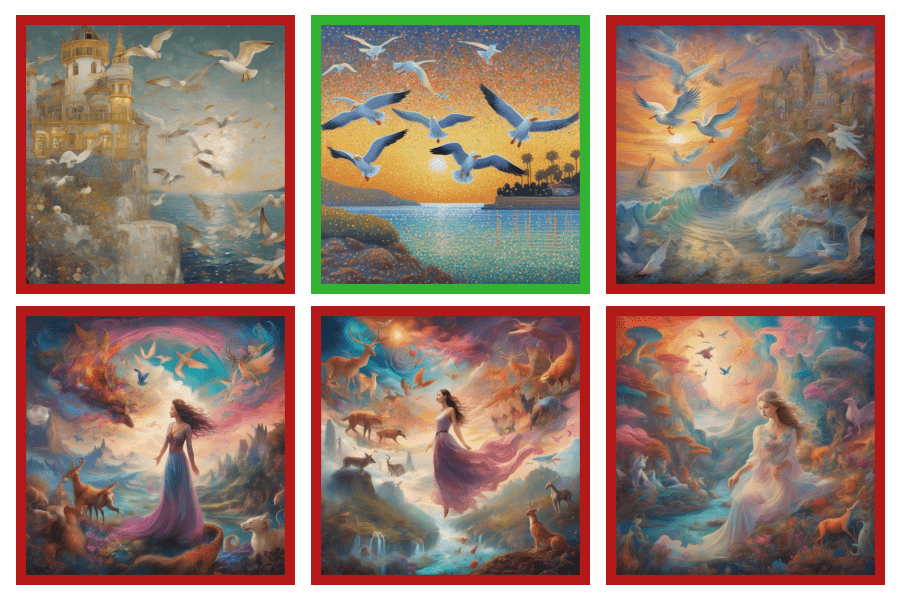}
        \caption{CLIP \cite{radford2021learning}}
        
    \end{subfigure}
    \caption{Image retrieval for style-based evaluation on \ourd Dataset, with images ranked highest to lowest from left to right. We show the ranked images for a fixed semantic for isolating stylistic variations. Green colors indicate \textcolor{correct}{correct} and red for \textcolor{wrong}{incorrect} retrievals.}
    \label{fig:synth_res_style}
\end{figure*}

\begin{figure*}[t!]
    \centering
    \begin{subfigure}[b]{0.1\textwidth}
        \centering
        \includegraphics[width=0.97\textwidth]{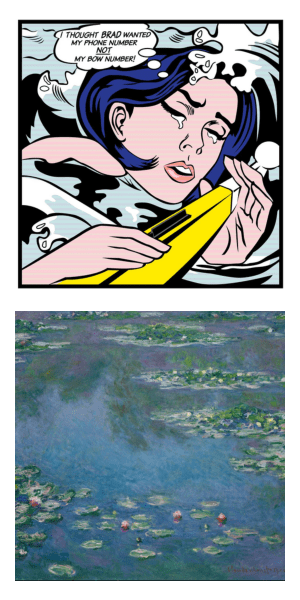}
        \caption{Queries}
        
    \end{subfigure}
    \hfill
    \begin{subfigure}[b]{0.29\textwidth}
        \centering
        \includegraphics[width=1\textwidth]{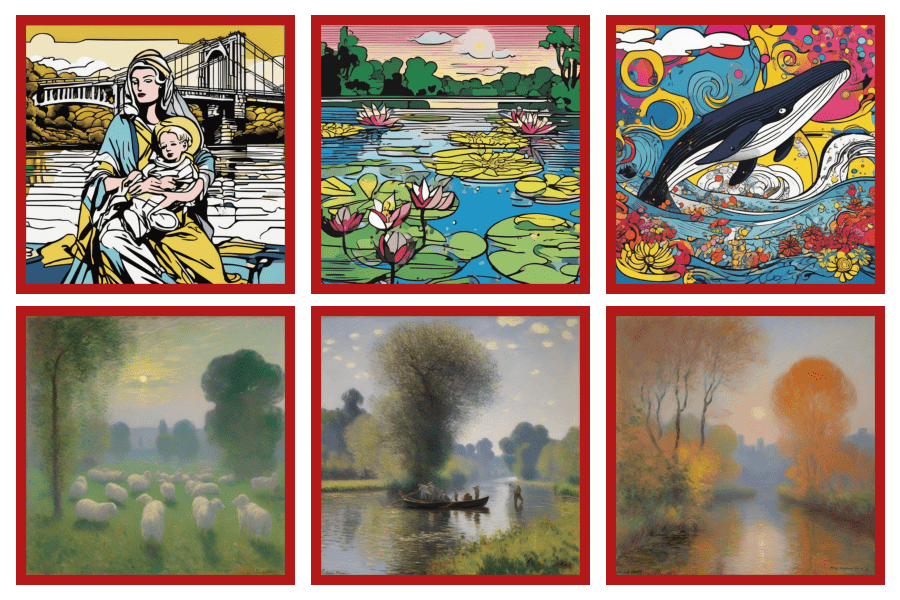}
        \caption{\ours (ours)}
        
    \end{subfigure}
    \hfill
    \begin{subfigure}[b]{0.29\textwidth}
        \centering
        \includegraphics[width=1\textwidth]{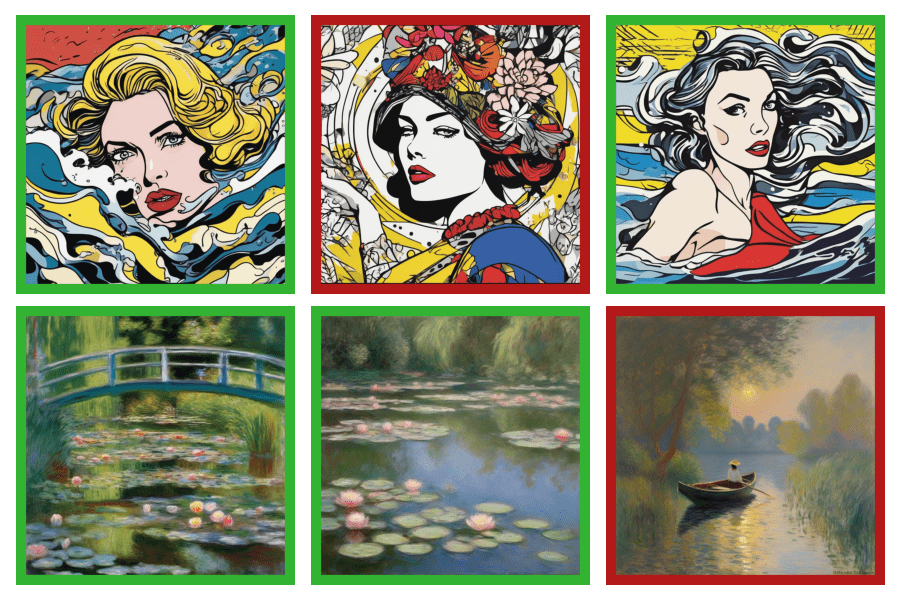}
        \caption{CSD \cite{somepalli2024measuring}}
        
    \end{subfigure}
    \hfill
    \begin{subfigure}[b]{0.29\textwidth}
        \centering
        \includegraphics[width=1\textwidth]{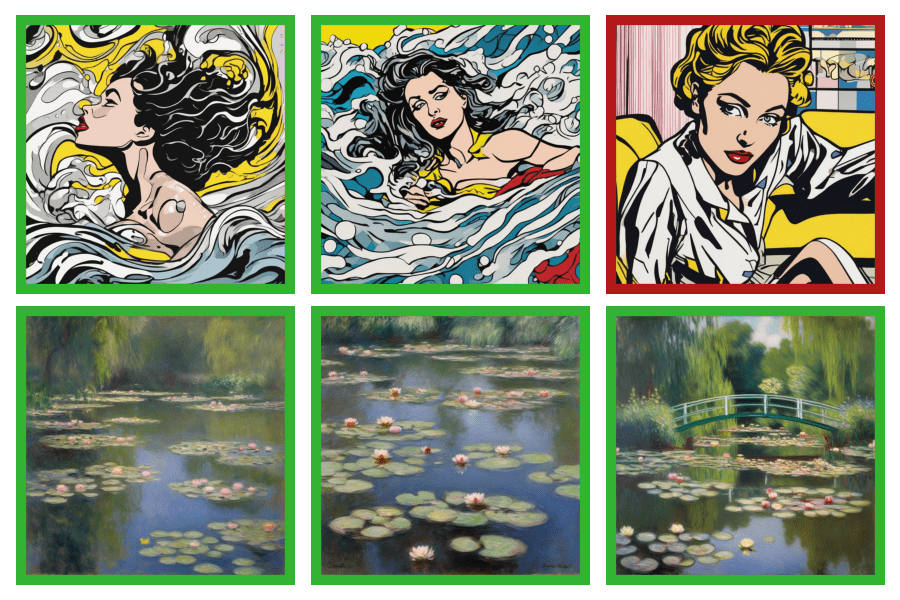}
        \caption{CLIP \cite{radford2021learning}}
        
    \end{subfigure}
    \caption{Image retrieval for semantic-based evaluation on \ourd Dataset, with images ranked highest to lowest from left to right. The results suggest that our retrieval emphasizes styles rather than semantic content. Green colors indicate \textcolor{correct}{correct} and red for \textcolor{wrong}{incorrect} retrievals.}
    \label{fig:synth_res_sem}
\end{figure*}

\section{Experiments}
\label{sec:exp}
We use the Stable Diffusion v2.1 model initialized with weights from Hugging Face. Introspective attribution is training-free; we extract feature tensors from the model to obtain the \ours image representation using an NVIDIA RTX 3090ti GPU with batch size 4. Each image is center-cropped to resolution $512 \times 512$. The guidance scale is irrelevant since we do not use a text prompt. 
% {\color{red}WHY ARE WE TALKING ABOUT THIS AND CFG?} 
Unless otherwise noted, \ours is implemented with $idx=1$, timestep $t = 25$, and the 2-Wasserstein distance.

\subsection{Datasets and Baselines}

\textbf{Dataset.}
\textit{DomainNet}~\cite{peng2019moment} comprises images from six domains: Clipart, Infograph, Painting, Quickdraw, Real, and Sketch, with approximately equal representation. Due to strong stylistic similarities between Quickdraw and Sketch domains, we excluded the former. The test set, consisting of $127,855$ images, was randomly divided into $25,571$ query images and $102,284$ reference images.
\textit{WikiArt}~\cite{saleh2015large} contains $80,096$ fine art images representing $1,119$ artists across $27$ genres. We randomly split this dataset into $64,090$ reference and $16,006$ query images. As proposed in~\cite{somepalli2024measuring}, each artist defines a separate class, correlating style and artist. WikiArt poses a greater style retrieval challenge since the large number of artists makes chance level matching performance $0.09\%$, significantly less than the $20\%$ of DomainNet.
These are the only two publicly available datasets for style retrieval\footnote{The Behance Artistic Media (BAM) dataset~\cite{ruta2021aladin} is not publicly available.}, making the proposed \ourd dataset a valuable addition to the field. We also conduct experiments on \ourd, which has $100$ queries, each ranking $60,000$ images.
% {\color{red} Are THESE THE ONLY AVAILABLE? WHAT ABOUT LAION AESTHETIC?}

\textbf{Evaluation Metrics.}
Style attribution performance is measured by two standard retrieval metrics~\cite{somepalli2024measuring, wang2023attribution}: Recall@$k$ and mean Average Precision (mAP)@$k$, for $k \in {1, 10, 100}$.

\textbf{Baselines.}
We compared \ours to several baselines from the style attribution and representation learning literature. For style attribution, we consider two recent methods, Contrastive Style Descriptors (CSD)~\cite{somepalli2024measuring} and Generative Data Attribution (GDA)~\cite{wang2023attribution}\footnote{We use the \texttt{real mapper} to obtain features}, which employ fine-tuning on additional datasets for enhanced performance.
This is complemented by a training-free method based on VGG Gram Matrices~\cite{gatys2016image}. We utilize the final layer embeddings as feature representations for all models under comparison. This standardized approach ensures a fair comparison across diverse architectures and training paradigms, allowing us to isolate and evaluate the effectiveness of each representation for style attribution.
% {\color{red} These should be described at a high level in the lit review, since they are main competitors}. 
For representation learning, we consider multiple vision foundational models, including CLIP~\cite{radford2021learning}, DINO~\cite{caron2021emerging}, MoCo~\cite{he2019momentum}, and SSCD~\cite{pizzi2022self}, which have proven effective in the literature for various computer vision tasks.

\begin{table}
  \centering
  \footnotesize
  \begin{tabular}{l | c c }
    \toprule
     % & \multicolumn{4} {c} {\ourd Dataset} \\
    % \midrule
    Method & Style-Eval@10 $\uparrow$ & Semantic-Eval@10 $\downarrow$\\
    \midrule
    DINO ViT-B/16 \cite{caron2021emerging} & \cellcolor{tabsecond}0.682 &  0.190 \\
    CLIP ViT-L \cite{radford2021learning}&  0.642 & 0.223 \\
    \midrule
    GDA DINO ViT-B \cite{wang2023attribution}  & 0.541 & \cellcolor{tabsecond}0.143 \\
    CSD ViT-L \cite{somepalli2024measuring} & 0.636& 0.183 \\
    \midrule
    % StyFeat-Idx: $0$, $t$: 25 & 0.866 & 0.830 & 0.474 & 0.866 & 0.879 & 0.911  & 0.928 & 0.923 & 0.778 & 0.928 & 0.972 & 0.996 \\
    \ours (Ours) & {\cellcolor{tabfirst} 0.823}& {\cellcolor{tabfirst} 0.114}\\ 
    % StyFeat-Idx: $2$, $t$: 25 & 0.888 & 0.852 & 0.521& 0.888 & 0.907 & 0.938 & 0.952 & 0.948 & 0.847 & 0.952 & 0.981 & 0.996  \\ 
    % StyFeat-Idx: $3$, $t$: 25 & 0.872 & 0.836 & 0.491& 0.872 & 0.886 & 0.916  & 0.935 & 0.934 & 0.812 & 0.935 & 0.974 & 0.994 \\ 
    \bottomrule
  \end{tabular}
  \caption{Evaluation results on the \ourd dataset.}
  \label{tab:synthetic}
\end{table}

\subsection{WikiArt and DomainNet Results}

Table~\ref{tab:main} shows that \ours significantly outperforms the various baselines for style attribution the WikiArt and DomainNet datasets.  Its gains are quite significant for the more practically relevant (lower) values of $k$. In the more challenging WikiArt dataset, \ours achieves a gain of close to $20$ points over the previous best approach (CSD) for both mAP and Recall when $k = 1$. For $k= 10$, the mAP gain is more than $30$ points. The improvement drops at $k =100$ because most artists have less than $100$ images, making it impossible to achieve a perfect mAP score and masking the differences between methods. In DomainNet, when compared to the baselines, \ours has gains of around $10\%$ for $k =1$ in both the metrics and a similar improvement on mAP@$10$. We note that these are gains of large magnitude, not commonly seen in retrieval applications, suggesting that introspective style attribution is vastly superior to methods based on external models. Fig.~\ref{fig:wikiart_res} illustrates this by showing examples of the three top matches retrieved by \ours, CSD, and GDA for the queries on the left. The middle row shows how the baselines are prone to retrieving images semantically similar to the query rather than of the same style, while \ours is more focused on style.

\subsection{\ourd Dataset Results}

% \textcolor{red}{explain: same semantic+varying style case, 1*12 in 50*12 are correct, then average 50 results?
% same style+varying semantics case, 1*12 in 100*12 are correct, then average 100 results? the best results would be random, which is 0.01?}

For the style-based evaluation, we calculate the rank of the $12 \times 100$ positives, averaged for the $100$ different query images, and semantic-based evaluation is done for $12 \times 50$ positives, averaged for the $100$ queries. The higher the score for style-based assessment, the better, and the lower, the better for semantic-based. If a model is truly agnostic to semantics and solely focuses on style, the semantic-based retrieval should be random and around $0.01$.

Table~\ref{tab:synthetic} compares different approaches on the \ourd dataset. In particular, \ours outperforms all the models in both evaluations by $20\%$ in style and $8\%$ in semantics compared to the second-best method, showing its effectiveness at isolating style. Fig.~\ref{fig:synth_res_style} shows some results of the style-based evaluation, illustrating how \ours has a significantly higher capacity than the baseline representation to focus on stylistic features. Our method ignores semantic details, as confirmed in Fig.~\ref{fig:synth_res_sem}, which refers to the semantics-based evaluation and shows that existing methods have a much stronger tendency to retrieve images of semantics similar to the query.

\subsection{Model Size}

Table~\ref{tab:param} compares model sizes across state-of-the-art methods. Our approach (\ours) has the highest parameter count primarily due to the large Stable Diffusion UNet, which represents a limitation of our method as it requires more substantial computational resources.

% Table \ref{tab:synthetic} compares different approaches on the \ourd dataset.  While the recall metric provides less information about the differences, with all of them being $1.00$, \ our precision is substantially better compared to other methods. In particular, \ours outperforms a strong baseline model, CLIP, by $3\%$ in the mAP@$5$ metric, showing that it is much more effective at retrieving images crafted by ``hacked" prompts that indirectly refer to a specific artistic style. This can also be seen by the visualizations of Fig. \ref{fig:synth_res}.

\subsection{Ablations}
In this section, we ablate various parameters and discuss the best choices for timestep $t$, block index $idx$, and similarity metric. All experiments are performed on WikiArt.

\begin{figure}[t]
  \centering
   \includegraphics[width=1.0\columnwidth]{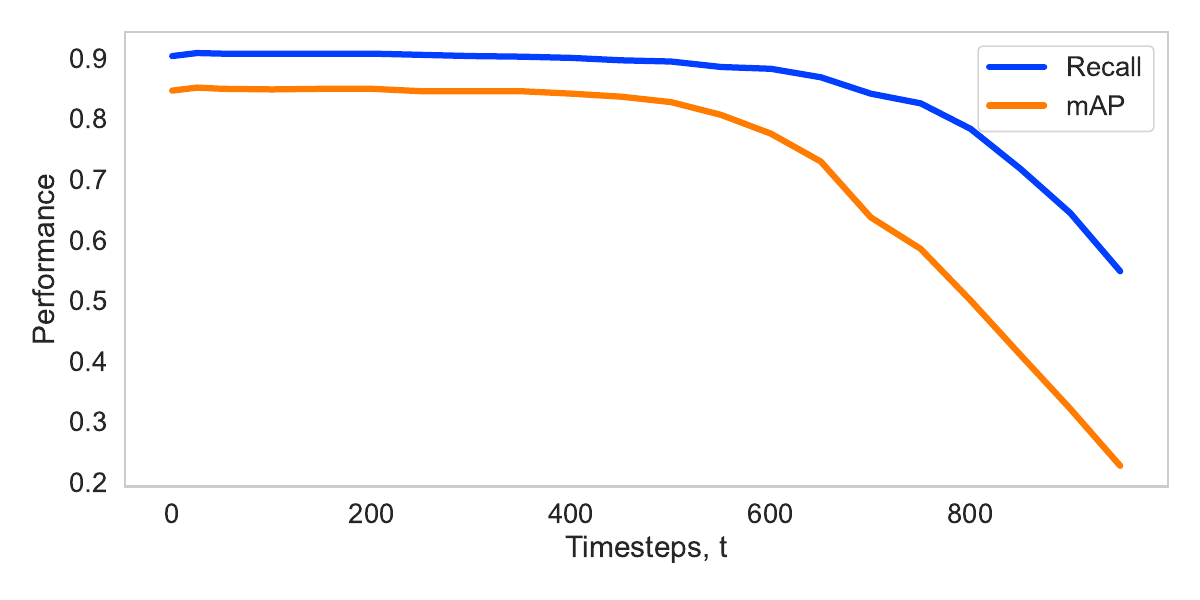}

   \caption{Effects of different timesteps on WikiArt image retrieval.}
   \label{fig:perf_time}
\end{figure}

\paragraph{Timestep.}
Fig. \ref{fig:perf_time} shows how style attribution performance (Recall and mAP at $k =10$) varies with the timestep $t \in [0, 950]$ for the implementation of \ours with features of up block index $idx = 1$ and the 2-Wasserstein distance. The performance remains stable until around $t = 400$, dropping significantly after that. We choose $t = 25$ as the default value for the timestep hyperparameter of \ours as this performed the best on WikiArt and DomainNet. Similar observations hold for up block indices $idx \in {0, 2, 3}$, which are presented in the supplementary.

\begin{table}
  \centering
  \footnotesize
  \begin{tabular}{l | c c c | c c c }
    \toprule
    % Block & \multicolumn{6} {c|} {WikiArt} \\
    % \midrule
     & \multicolumn{3}{c|}{mAP@$k$} & \multicolumn{3}{c}{Recall@$k$}\\
    Block & 1 & 10 & 100 & 1 & 10 & 100  \\
    \midrule
    $idx=0$ & 0.866 & 0.830 & 0.474 & 0.866 & 0.879 & 0.911 \\
     $idx=1$ & 0.887 & {\cellcolor{tabfirst} 0.852} & {\cellcolor{tabfirst} 0.525}& 0.887 & {\cellcolor{tabfirst} 0.909} & {\cellcolor{tabfirst} 0.941} \\ 
    $idx=2$& {\cellcolor{tabfirst} 0.888} & {\cellcolor{tabfirst} 0.852} & 0.521& {\cellcolor{tabfirst} 0.888} & 0.907 & 0.938 \\ 
    $idx=3$ & 0.872 & 0.836 & 0.491& 0.872 & 0.886 & 0.916 \\ 
    \bottomrule
  \end{tabular}
  \caption{Ablation on feature block index selection, all with $t = 25$.}
  \label{tab:up_idx}
\end{table}

\paragraph{Up block index.}  Table \ref{tab:up_idx} compares the performance of \ours with $t = 25$ and the 2-Wasserstein distance for different values of the 
up block index $idx$. The best overall performance on both datasets is achieved when $idx = 1$, but \ours is robust to the choice of features used to create the feature representation. We compare the model size for each $idx$ in supplementary.
% Further, by increasing the $idx$ value, we increase the size of our model significantly, as shown in Table \ref{tab:param_ablate}. Therefore, we set $idx=1$ as the default value for the $idx$ hyperparameter to get the best results and minimal model size.

% \begin{table}[t]
%   \centering
%   \footnotesize
%   \begin{tabular}{l | c c c | c c c}
%     \toprule
%     Metric & \multicolumn{6} {c} {WikiArt} \\
%     \midrule
%     & \multicolumn{3}{c|}{mAP@$k$} & \multicolumn{3}{c}{Recall@$k$}\\
%     & 1 & 10 & 100 & 1 & 10 & 100\\
%     \midrule
%     Euclidean & {\bf 0.888} & {\bf 0.852} & 0.521 & {\bf 0.888} & 0.907 & 0.938\\
%     Gram   & 0.869 & 0.835 & 0.491 & 0.869 & 0.885 & 0.916\\
%     JSD & {\bf 0.888} & 0.849 & {\bf 0.525} & {\bf 0.888} & 0.908 & 0.940\\
%     2-Wasserstein & 0.887 & {\bf 0.852} & {\bf 0.525} & 0.887 & {\bf 0.909} & {\bf 0.941}\\
%     \bottomrule
%   \end{tabular}
%   \caption{Finding the best metric}
%   \label{tab:ablation}
% \end{table}
\begin{table}[t]
  \centering
  \footnotesize
  \begin{tabular}{l | c c c | c c c}
    \toprule
     % & \multicolumn{6} {c} {WikiArt} \\
    % \midrule
    & \multicolumn{3}{c|}{mAP@$k$} & \multicolumn{3}{c}{Recall@$k$}\\
    Metric & 1 & 10 & 100 & 1 & 10 & 100\\
    \midrule
    $L_2$ & {\cellcolor{tabfirst} 0.888} & {\cellcolor{tabfirst} 0.852} & 0.521 & {\cellcolor{tabfirst} 0.888} & 0.907 & 0.938\\
    Gram   & 0.869 & 0.835 & 0.491 & 0.869 & 0.885 & 0.916\\
    JSD & {\cellcolor{tabfirst} 0.888} & 0.849 & {\cellcolor{tabfirst} 0.525} & {\cellcolor{tabfirst} 0.888} & 0.908 & 0.940\\
    $W_2$ & 0.887 & {\cellcolor{tabfirst} 0.852} & {\cellcolor{tabfirst} 0.525} & 0.887 & {\cellcolor{tabfirst} 0.909} & {\cellcolor{tabfirst} 0.941}\\
    \bottomrule
  \end{tabular}
  \caption{Comparison on different metrics.}
  \label{tab:ablation}
\end{table}

% \begin{table}
%   \centering
%   \footnotesize
%   \begin{tabular}{l | c }
%     \toprule
%     Method & No. of Parameters \\
%     \midrule
%     % DINO ViT-B/16 \cite{caron2021emerging} & 21M\\
%     % CLIP ViT-L \cite{radford2021learning} & 427M\\
%     % %     DINO ViT-B/16 \cite{caron2021emerging} & 21,665,664 (21M)\\
%     % % CLIP ViT-L \cite{radford2021learning} & 427,616,513 (427M)\\
%     % \midrule
%     % GDA DINO ViT-B \cite{wang2023attribution} & 86M \\
%     % CSD ViT-L \cite{somepalli2024measuring} &  305M\\
%     % % GDA DINO ViT-B \cite{wang2023attribution} & 86,389,248 (86M) \\
%     % % CSD ViT-L \cite{somepalli2024measuring} &  304,752,640 (305M)\\
%     % \midrule
%     % \ours ($idx$ = $0$) & 548,055,112 (548M)\\ 
%     % \ours ($idx$ = $1$) & 808,352,072 (808M) \\ 
%     % \ours ($idx$ = $2$) & 880,748,232 (881M)\\ 
%     % \ours ($idx$ = $3$) & 900,050,312 (900M)\\ 
%     \ours ($idx$ = $0$) & 548M\\ 
%     \ours ($idx$ = $1$) & 808M \\ 
%     \ours ($idx$ = $2$) & 881M\\ 
%     \ours ($idx$ = $3$) & 900M\\ 
%     \bottomrule
%   \end{tabular}
%   \caption{Comparison of the Parameter Count for different up block indices, $idx$ of our model.}
%   \label{tab:param_ablate}
% \end{table}

\paragraph{Similarity metric.} Table \ref{tab:ablation} compares the performance of \ours for the different similarity metrics of Section~\ref{sec:method}, 
on WikiArt, using $t = 25$ and $idx = 1$. The 2-Wasserstein distance has the best overall performance, albeit it is only marginally better than the Euclidean Distance and the JSD. This shows that \ours has a fairly stable performance with respect to the choice of similarity measure. We choose the 2-Wasserstein distance as the default metric for implementing \ours. 

\begin{table}
  \centering
  \footnotesize
  \begin{tabular}{l | c }
    \toprule
    Method & Number of Parameters \\
    \midrule
    DINO ViT-B/16 \cite{caron2021emerging} & 21M\\
    CLIP ViT-L \cite{radford2021learning} & 427M\\
    %     DINO ViT-B/16 \cite{caron2021emerging} & 21,665,664 (21M)\\
    % CLIP ViT-L \cite{radford2021learning} & 427,616,513 (427M)\\
    \midrule
    GDA DINO ViT-B \cite{wang2023attribution} & 86M \\
    CSD ViT-L \cite{somepalli2024measuring} &  305M\\
    % GDA DINO ViT-B \cite{wang2023attribution} & 86,389,248 (86M) \\
    % CSD ViT-L \cite{somepalli2024measuring} &  304,752,640 (305M)\\
    \midrule
    % \ours ($idx$ = $0$) & 548,055,112 (548M)\\ 
    % \ours ($idx$ = $1$) & 808,352,072 (808M) \\ 
    % \ours ($idx$ = $2$) & 880,748,232 (881M)\\ 
    % \ours ($idx$ = $3$) & 900,050,312 (900M)\\ 
    % \ours ($idx$ = $0$) & 548M\\ 
    \ours ($idx$ = $1$) & 808M \\ 
    % \ours ($idx$ = $2$) & 881M\\ 
    % \ours ($idx$ = $3$) & 900M\\ 
    \bottomrule
  \end{tabular}
  \caption{Comparison on the model sizes.} 
  % {\color{red}THIS NEEDS TO BE DISCUSSED IN THE  TEXT, NOT ONLY CONCLUSIONS.}}
  \label{tab:param}
\end{table}

\section{Discussion}

We presented \ours, a training-free style attribution model that extracts stylistic features to compute similarity scores between images. \ours was shown to outperform existing attribution methods across all metrics. To address the lack of datasets to evaluate style attribution, we developed \ourd, a synthetic dataset that isolates stylistic elements and enables fine-grained attribution evaluation. The robust performance of \ours on both this and previously used datasets establishes it as a state-of-the-art method for style attribution, with potential applications in protecting artists' intellectual property rights.

\ours has $2.5\times$ more parameters than CSD (Table \ref{tab:param}), which constrains its deployment on memory-limited devices. However, it can leverage the features produced by a diffusion model, without additional memory overhead, for style attribution of the images generated by that model. While our current evaluation focuses on Stable Diffusion v2.1, future work could explore compatibility with other pre-trained diffusion models, investigate performance gains through additional linear/MLP layers, and examine applications in diffusion-based style transfer.

\vspace{1em}
{\footnotesize \textbf{Acknowledgements.}~This work was partially funded by NSF awards IIS-2303153 and NAIRR-240300, a NVIDIA Academic grant, and a gift from Qualcomm. We also acknowledge and thank the use of the Nautilus platform for some of the experiments.}

% \section{Acknowledgments}
% We acknowledge and thank NRP Nautilus Portal for providing the compute resources to run our experiments.

{
    \small
    \bibliographystyle{ieeenat_fullname}
    \bibliography{main}
}
% \newpage
\newpage
\appendix
\twocolumn[
\centering
\Large
\vspace{0.5em}Supplementary Material \\
\vspace{1.0em}
]
% \clearpage
% \setcounter{page}{1}
% \title{\ours: Training-Free Introspective Style Attribution \\ using Diffusion Features}
% \maketitlesupplementary
\section{Overview}

In this supplementary material, we present further details about our methodology and experimental findings. Specifically, we provide an analysis of the hyper-parameter selection for \ours features in Sections~\ref{sec:metrics} and~\ref{sec:ablations}. 
% We conduct a detailed examination of the t-SNE distribution patterns in Section~\ref{sec:tsne}
Furthermore, we elaborate on our prompt engineering process utilizing ChatGPT for style isolation in the synthesis of \ourd dataset in Section~\ref{sec:proompt}. Finally, we present additional experimental results and analyses on both the WikiArt and \ourd datasets in Section~\ref{sec:results}. Our codes and \ourd dataset will be released.

\section{Similarity Metrics}\label{sec:metrics}
In this section, we include the details and formulate the Euclidean, Gram Matrices (using \ours features), and Jensen-Shannon Divergence (JSD) metrics as discussed in section 3.3 of the main text.

The $L_2$ distance (Euclidean distance) 
\begin{equation}
    L_2(\mu_1, \mu_2)^2 = \|\mu_1 - \mu_2\|_2^2,
\end{equation}
ignores covariance information and does not have an interpretation as a measure of similarity between probability distributions. This is also the case for the Frobenius norm between the Gram matrices, which is popular in the style transfer literature. It extracts deep features from the two images and then takes the outer product of their corresponding mean vectors $\mu_1$ and $\mu_2$, 
\begin{equation}
    \text{Gram}(\mu_1, \mu_2) = \|\mu_1 \mu_1^T - \mu_2\mu_2^T\|_F.
\end{equation}
Another popular similarity measure between distributions is the Kullback–Leibler (KL) divergence. For two multivariate Gaussians, it takes the form
\begin{small}
\begin{eqnarray}
\lefteqn{\mathrm{KL}((\mu_1, \Sigma_1) || (\mu_2, \Sigma_2)) =} &&\\
&& \frac{1}{2} \left[ \log \frac{|\Sigma_2|}{|\Sigma_1|} + \mathrm{tr}(\Sigma_2^{-1}\Sigma_1) + (\mu_2 - \mu_1)^T \Sigma_2^{-1} (\mu_2 - \mu_1) \right] \nonumber.
\end{eqnarray}
Note that the KL divergence is not symmetric. To address this, the Jensen-Shannon Divergence (JSD) averages the KL divergence going in both directions, 
\begin{equation}
    \mathrm{JSD}(I_1 || I_2) = \frac{1}{2}\mathrm{KL}(I_1 || I_2) + \frac{1}{2}\mathrm{KL}(I_2 || I_1).
\end{equation}
\end{small}

\section{Timestep and Block Index}\label{sec:ablations}

% We perform sweep over the timesteps to analyze the performance of \our features for all up-block indices $idx$ on the WikiArt Dataset, and the results are shown in Figure \ref{fig:map} and \ref{fig:recall}. The up-block indices $1$ and $2$ have the best results for mAP @ $10$ and recall @ $10$. From these, we choose $idx = 1$ as this has a larger channel size ($C$) and lower network parameters, as shown in Table \ref{tab:param}.

\begin{figure}
    \centering
    \includegraphics[width=0.9\linewidth]{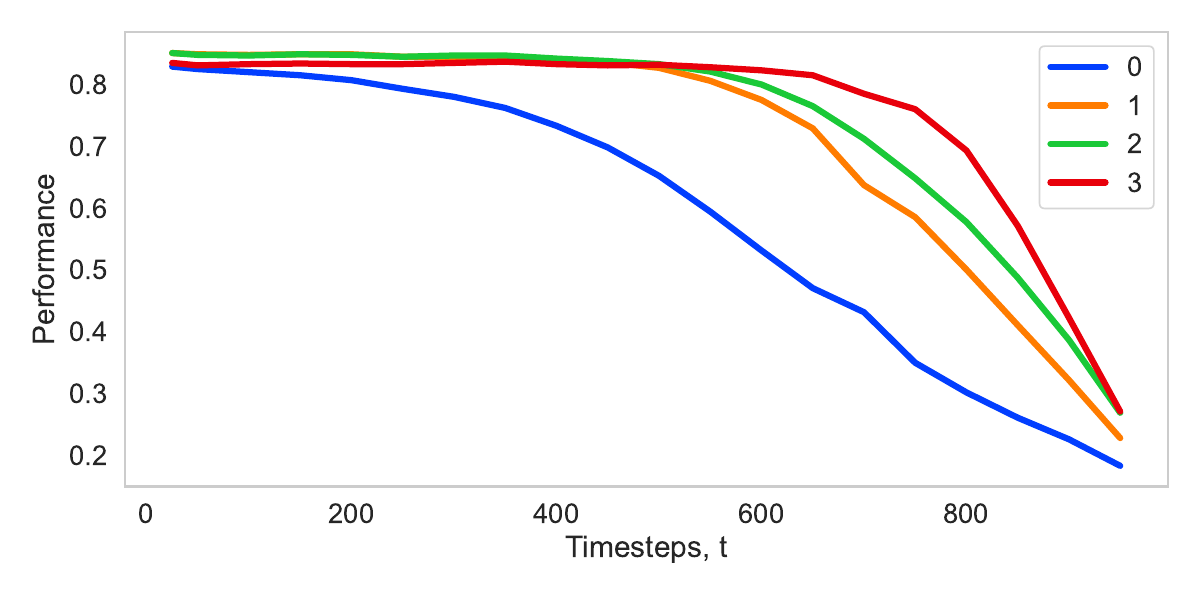}
    \caption{Precision (mAP@10) as a function of timestep $t$ for different up-block indices ($idx$) on the WikiArt dataset.}
    \label{fig:map}
\end{figure}

\begin{figure}
    \centering
    \includegraphics[width=0.9\linewidth]{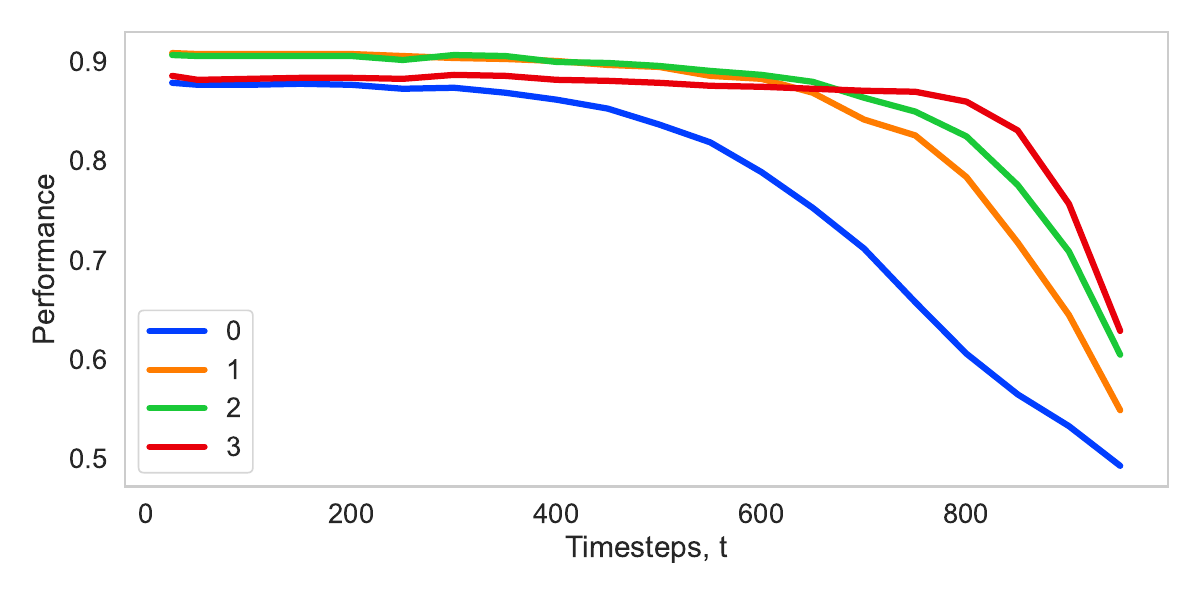}
    \caption{Recall (Recall@10 ) as a function of timestep $t$ for different up-block indices ($idx$) on the WikiArt dataset.}
    \label{fig:recall}
\end{figure}

We study the choices of timestep $t$ and block indices $idx$ of \ours by evaluating the image retrieval performance on the WikiArt Dataset. The results are presented in Figs.~\ref{fig:map} and~\ref{fig:recall}, showing that best performance is obtained with $t < 500$ and up-block indices $idx=1$ or $idx=2$, achieving a balanced trade-off between mAP@$10$ and recall@$10$.

\begin{table}
  \centering
  \footnotesize
  \begin{tabular}{l | c c c | c c c }
    \toprule
    % Block & \multicolumn{6} {c|} {WikiArt} \\
    % \midrule
     & \multicolumn{3}{c|}{mAP@$k$} & \multicolumn{3}{c}{Recall@$k$}\\
    Block & 1 & 10 & 100 & 1 & 10 & 100  \\
    \midrule
    $idx=0$ & 0.947 & 0.925 & 0.791 & 0.947 & 0.970 & 0.991 \\
     $idx=1$ & {\cellcolor{tabfirst}0.954} & {\cellcolor{tabfirst} 0.949} & {\cellcolor{tabfirst} 0.850}& {\cellcolor{tabfirst}0.954} & {\cellcolor{tabfirst} 0.982} & {\cellcolor{tabfirst} 0.995} \\ 
    $idx=2$& 0.950 & 0.946 & 0.820& 0.950 & 0.964 & 0.987 \\ 
    $idx=3$ & 0.941 & 0.939 & 0.823 & 0.941 & 0.978 & 0.984 \\ 
    \bottomrule
  \end{tabular}
  \caption{DomainNet: Feature block index selection for $t = 25$.}
  \label{tab:up_idx_domain}
\end{table}

\begin{table}[t]
  \centering
  \footnotesize
  \begin{tabular}{l | c c c | c c c}
    \toprule
     % & \multicolumn{6} {c} {WikiArt} \\
    % \midrule
    & \multicolumn{3}{c|}{mAP@$k$} & \multicolumn{3}{c}{Recall@$k$}\\
    Metric & 1 & 10 & 100 & 1 & 10 & 100\\
    \midrule
    $L_2$ & {\cellcolor{tabfirst} 0.948} & 0.944 & 0.839 & {\cellcolor{tabfirst} 0.948} & 0.961 & 0.980\\
    $W_2$ & 0.954 & {\cellcolor{tabfirst} 0.949} & {\cellcolor{tabfirst} 0.850}& 0.954 & {\cellcolor{tabfirst} 0.982} & {\cellcolor{tabfirst} 0.995} \\
    \bottomrule
  \end{tabular}
  \caption{DomainNet: Comparison on different metrics.}
  \label{tab:ablation_domain}
\end{table}

On DomainNet, we observe consistent trends: 2-Wasserstein ($W_2$) distance improved mAP@10 by 0.05 compared to $L_2$ (Fig.~\ref{tab:up_idx_domain}),  features from block 1 led to a 0.03 improvement (Fig.~\ref{tab:ablation_domain}), time-step vs performance curve is similar. For variance, varying seeds we obtained 0.002/0.001 for mAP/Recall @10 respectively.

We chose $\text{idx} = 1$ as our default configuration for computational efficiency, as shown in Table~\ref{tab:param_sup}.

\begin{table}
  \centering
  \footnotesize
  \begin{tabular}{l | c | c}
    \toprule
    Method & Parameters & Channel ($C$)\\
    \midrule
    % \ours ($idx$ = $0$) & 548,055,112 (548M)\\ 
    % \ours ($idx$ = $1$) & 808,352,072 (808M) \\ 
    % \ours ($idx$ = $2$) & 880,748,232 (881M)\\ 
    % \ours ($idx$ = $3$) & 900,050,312 (900M)\\ 
    \ours ($idx$ = $0$) & 548M& 1280 \\ 
    \ours ($idx$ = $1$) & 808M & 1280\\ 
    \ours ($idx$ = $2$) & 881M& 640\\ 
    \ours ($idx$ = $3$) & 900M& 320\\ 
    \bottomrule
  \end{tabular}
  \caption{Comparison of the model size needed to compute \ours features and channel size for different choices of the up-block index ($idx$).}
  \label{tab:param_sup}
\end{table}

\section{Generating Prompts for \ourd dataset}\label{sec:proompt}

As explained in the main text, we curated a comprehensive collection of prompt-image pairs representing $50$ influential artists across various artistic movements and periods from the LAION Aesthetics Dataset: \texttt{Albert Bierstadt, Alex Gray, Alphonse Mucha, Amedeo Modigliani, Antoine Blanchard, Arkhip Kuindzhi, Carne Griffiths, Claude Monet, Cy Twombly, Diego Rivera, Edmund Dulac, Edward Hopper, Francis Picabia, Frank Auerbach, Frida Kahlo, George Seurat, George Stubbs, Gustav Klimt, Gustave Dore, Harry Clark, Hubert Robert, Ilya Repin, Isaac Levitan, Jamie Wyeth, Jan Matejko, Jan Van Eyck, John Atkinson Grimshaw, John Collier, John William Waterhouse, Josephine Wall, Katsushika Hokusai, Leonid Afremov, Lucian Freud, M.C. Escher, Man Ray, Mark Rothko, Paul Klee, Peter Paul Rubens, Picasso, RenÃ© Magritte, Richard Hamilton, Robert Delaunay, Roy Lichtenstein, Takashi Murakami, Thomas Cole, Thomas Kinkade, Vincent Van Gogh, Wassily Kandinsky, William Turner, Winslow Homer}. Using their seminal works as reference queries, we implemented a systematic prompt-generation strategy. For each artist, we derived a ``style" prompt and used $2$ of their paintings to generate ``semantic" prompts, subsequently employing a Stable Diffusion v2.1 to synthesize $12$ images per combination. This methodological approach yielded a richly diverse reference dataset comprising $60{,}000$ images ($50$ artistic styles $\times$ $12$ prompts $\times$ $100$ semantic descriptions).

We leveraged the ChatGPT to generate both style and semantic prompts systematically. 

\textbf{Style Prompt}. To create descriptions that effectively described the artistic style, we crafted a base system prompt as follows:

\begin{quote}
\textit{``You are a prompt generator for Stable Diffusion 2.1, and you are tasked with generating the style description of an artist.
\\
KEEP EVERYTHING SFW (SAFE FOR WORK).
\\
Only print the prompt without any other information and each prompt should not be more than 25 words.
''}
\end{quote}

\begin{figure}[t]
\begin{tcolorbox}[
       colback=blue!10,
       colframe=blue!50!black,
       title=User,
       width=8cm,
       halign title=flush right
   ]
   Create a prompt to generate an image in Vincent Van Gogh's style and include the artist's name.
\end{tcolorbox}
    \begin{tcolorbox}[
        colback=green!10,
        colframe=green!50!black,
        title={$\vcenter{\hbox{\includegraphics[width=4mm]{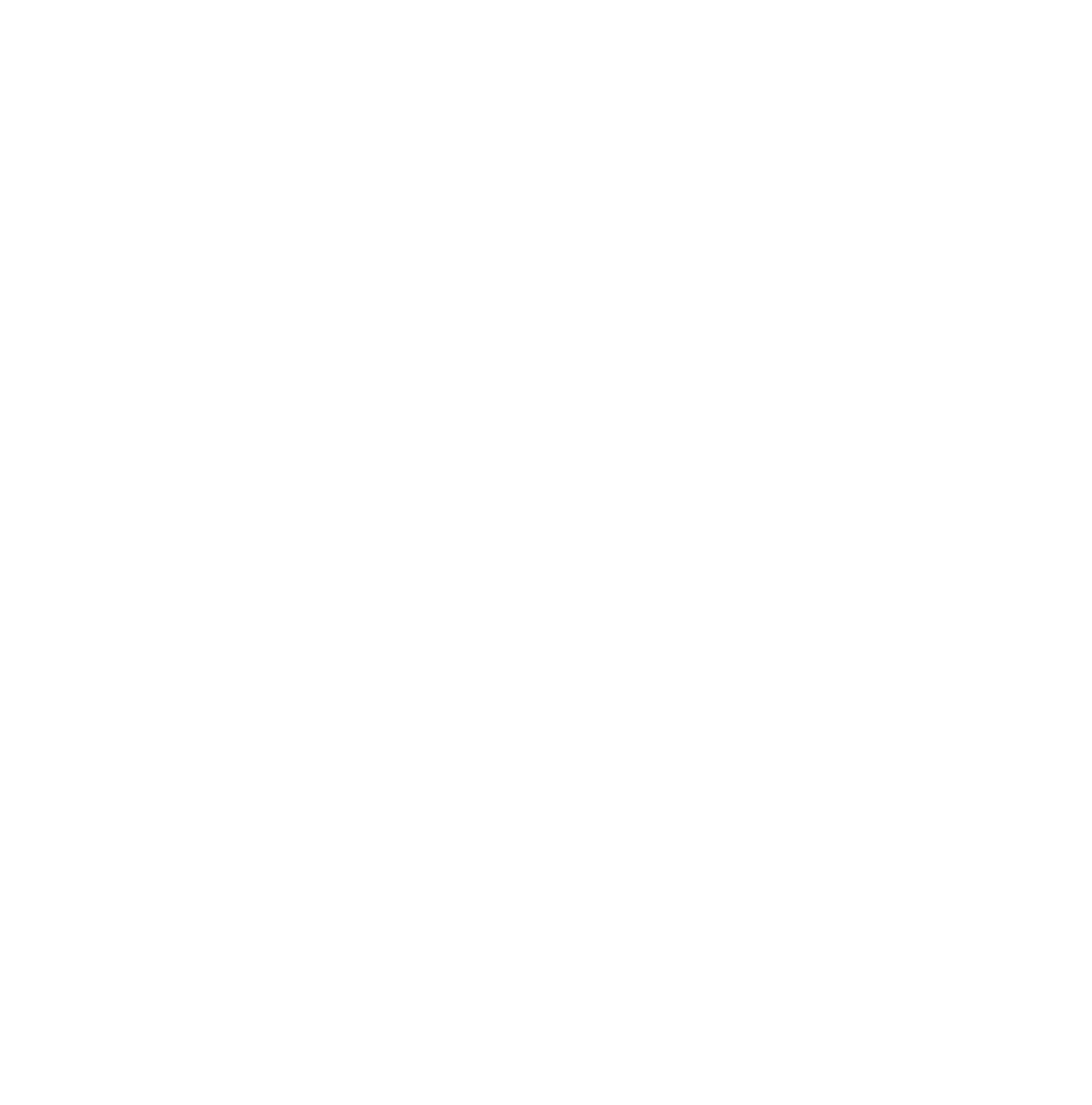}}}$ ChatGPT},
        width=8cm,
    ]

Van Gogh-inspired painting, swirling brushstrokes, vivid warm colors, emotive expression, night sky, Starry Night atmosphere, Vincent Van Gogh style.
\end{tcolorbox}
\caption{Style prompt generation using ChatGPT.}
\label{fig:style_gpt}
\end{figure}

The user prompt template is as follows:
\begin{quote}
\textit{``Create a prompt to generate an image in the style of [artist] and also include the artist's name''}
\end{quote}

with the variable inputs based on the chosen artist, with \texttt{artist} being the variable inputs. An example response by ChatGPT, for \texttt{Vincent Van Gogh}, is shown in Fig.~\ref{fig:style_gpt}.

% \textbf{Hacked Prompt}. To create descriptions that effectively mask the original artistic attribution while ensuring consistency and reproducibility in prompt generation, we crafted a base template query structured as follows:

% \begin{quote}
% \textit{``Can you give prompts to obtain a painting of [artist]'s [painting] without mentioning their name or their artwork in the prompt?''}
% \end{quote}

% with \texttt{artist} and \texttt{painting} being the variable inputs based on the chosen paintings. An example response by ChatGPT, for \texttt{Vincent Van Gogh} and his work \texttt{Starry Night}, is shown in Figure ~\ref{fig:gpt}.

\textbf{Semantic Prompt}. The semantic prompts are created for a given artist's painting by providing descriptions of the contents of the image without any style information. The base system prompt is as follows:

\begin{quote}
\textit{``You are a prompt generator for Stable Diffusion 2.1, and you are tasked with generating a prompt with the semantics of a given painting.\\
DO NOT HAVE ANY DESCRIPTION OF THE STYLE. KEEP EVERYTHING SFW (SAFE FOR WORK)\\
Only print the prompt without any other information and each prompt should not be more than 25 words.''}
\end{quote}

And the user prompt template is as follows:
\begin{quote}
\textit{``Create a prompt to generate an image with the semantics of {artists}'s {painting}''}
\end{quote}

\begin{figure}[t]
\begin{tcolorbox}[
       colback=blue!10,
       colframe=blue!50!black,
       title=User,
       width=8cm,
       halign title=flush right
   ]
  Create a prompt to generate an image with the semantics of Vincent Van Gogh's Starry Night.
\end{tcolorbox}
    \begin{tcolorbox}[
        colback=green!10,
        colframe=green!50!black,
        title={$\vcenter{\hbox{\includegraphics[width=4mm]{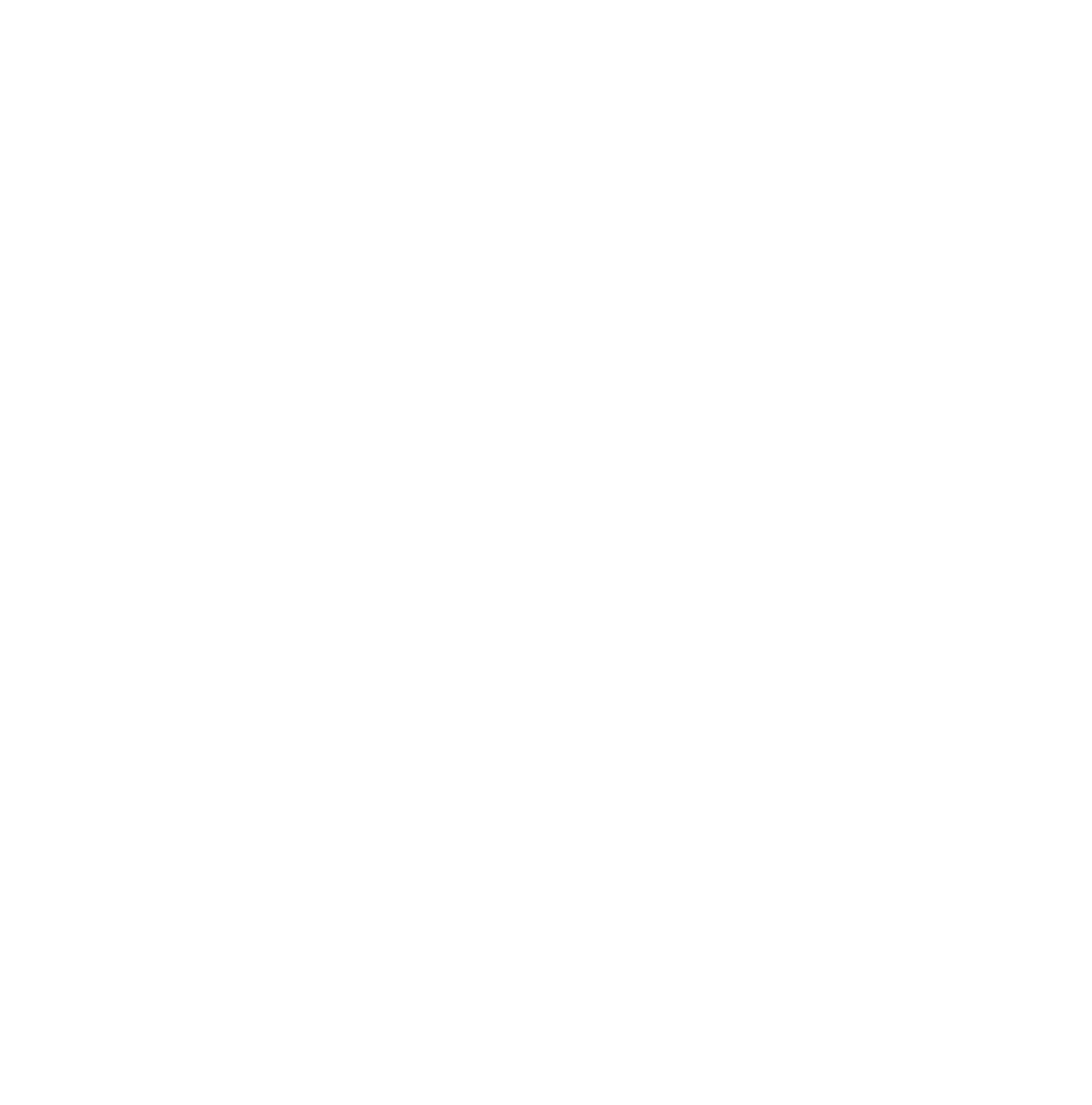}}}$ ChatGPT},
        width=8cm,
    ]

A night sky filled with vibrant stars above a quiet, small village nestled beneath rolling hills.
\end{tcolorbox}
\caption{Semantic prompt generation using ChatGPT.}
\label{fig:sem_gpt}
\end{figure}

With the variable inputs based on the chosen artist's painting, with \texttt{artist} and \texttt{painting} being the chosen artist's painting. An example response by ChatGPT, for \texttt{Vincent Van Gogh}'s \texttt{Starry Night}, is shown in Fig.~\ref{fig:sem_gpt}.

% \begin{quote}
% \textit{``Paint a dynamic nocturnal landscape featuring a turbulent sky filled with spiraling clouds and radiant stars. Depict a quiet town nestled beneath rolling hills, with cypress trees in the foreground."}
% \end{quote}
% is converted into an innocent prompt,
% \begin{quote}
% \textit{``Paint a nocturnal landscape featuring a sky filled with clouds and stars. Depict a quiet town nestled beneath rolling hills, with cypress trees in the foreground."} 
% \end{quote}

% In Figure~\ref{fig:synth_sample_sup}, examples of hacked and innocent prompts for the paintings shown on the left and the resulting synthesized images. Note how the prompts isolate style in various cases. 
% {\color{red} It seems to me that the bottom one is not a great example, because the kacked prompts do not seem to produce images in the style of the original on the left. Don't we have anything better (like the one at the top?}
    
\section{Results on WikiArt and \ourd Datasets}\label{sec:results}

\begin{figure*}[t!]
    \centering
    \begin{subfigure}[b]{0.128\textwidth}
        \centering
        \includegraphics[width=1\textwidth]{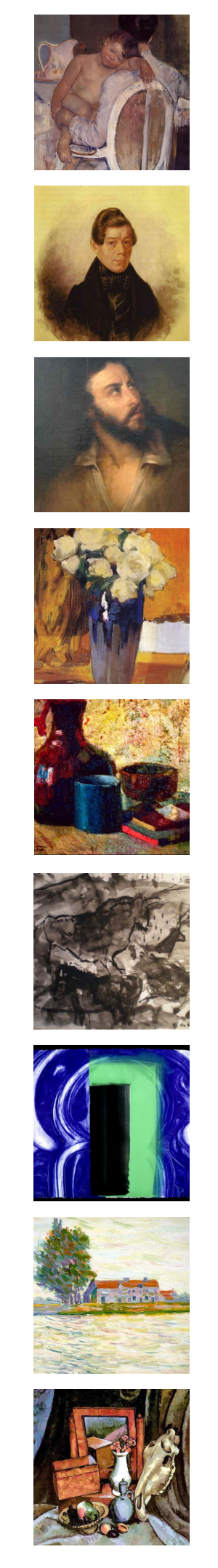}
        \caption{Queries}
        
    \end{subfigure}
    \hfill
    \begin{subfigure}[b]{0.42\textwidth}
        \centering
        \includegraphics[width=1\textwidth]{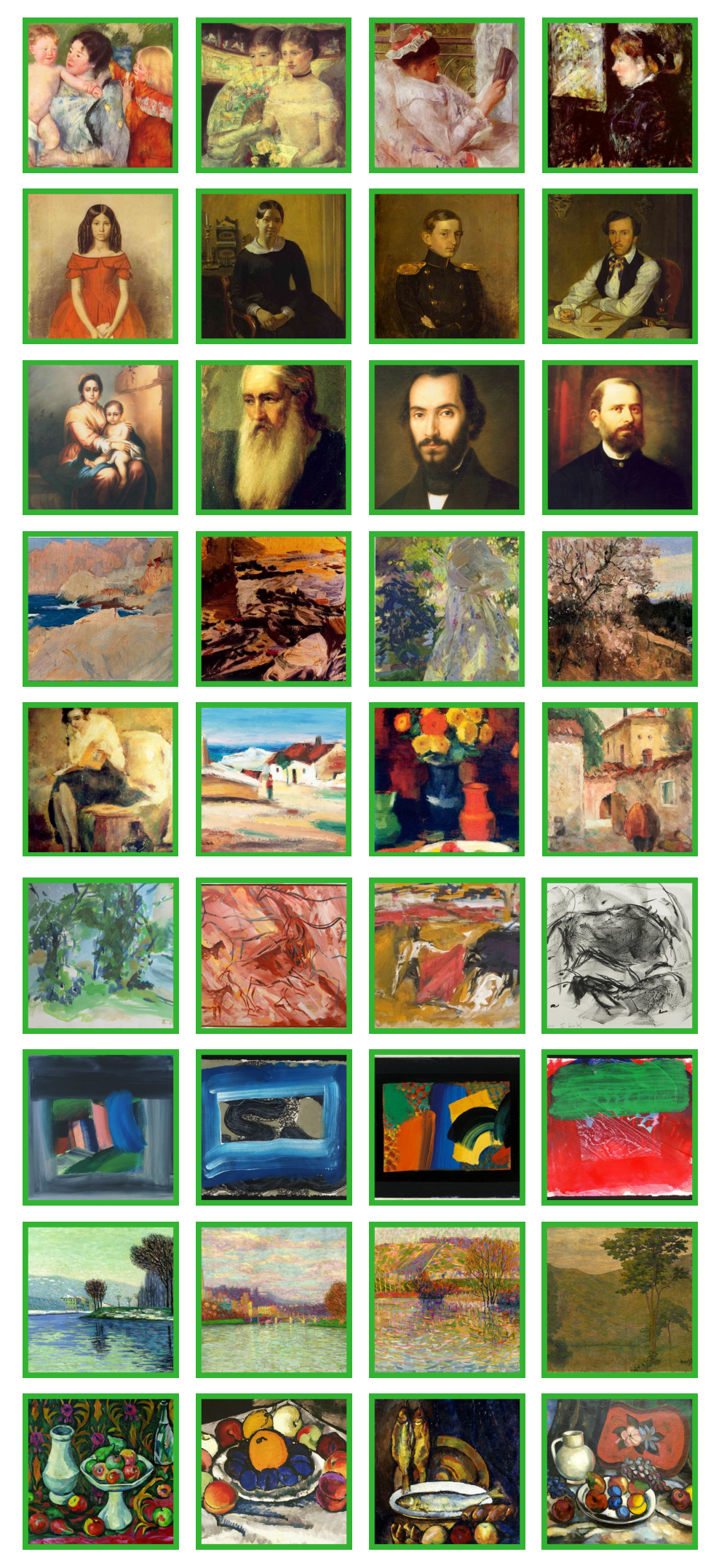}
        \caption{\ours (ours)}
        
    \end{subfigure}
    \hfill
    \begin{subfigure}[b]{0.42\textwidth}
        \centering
        \includegraphics[width=1\textwidth]{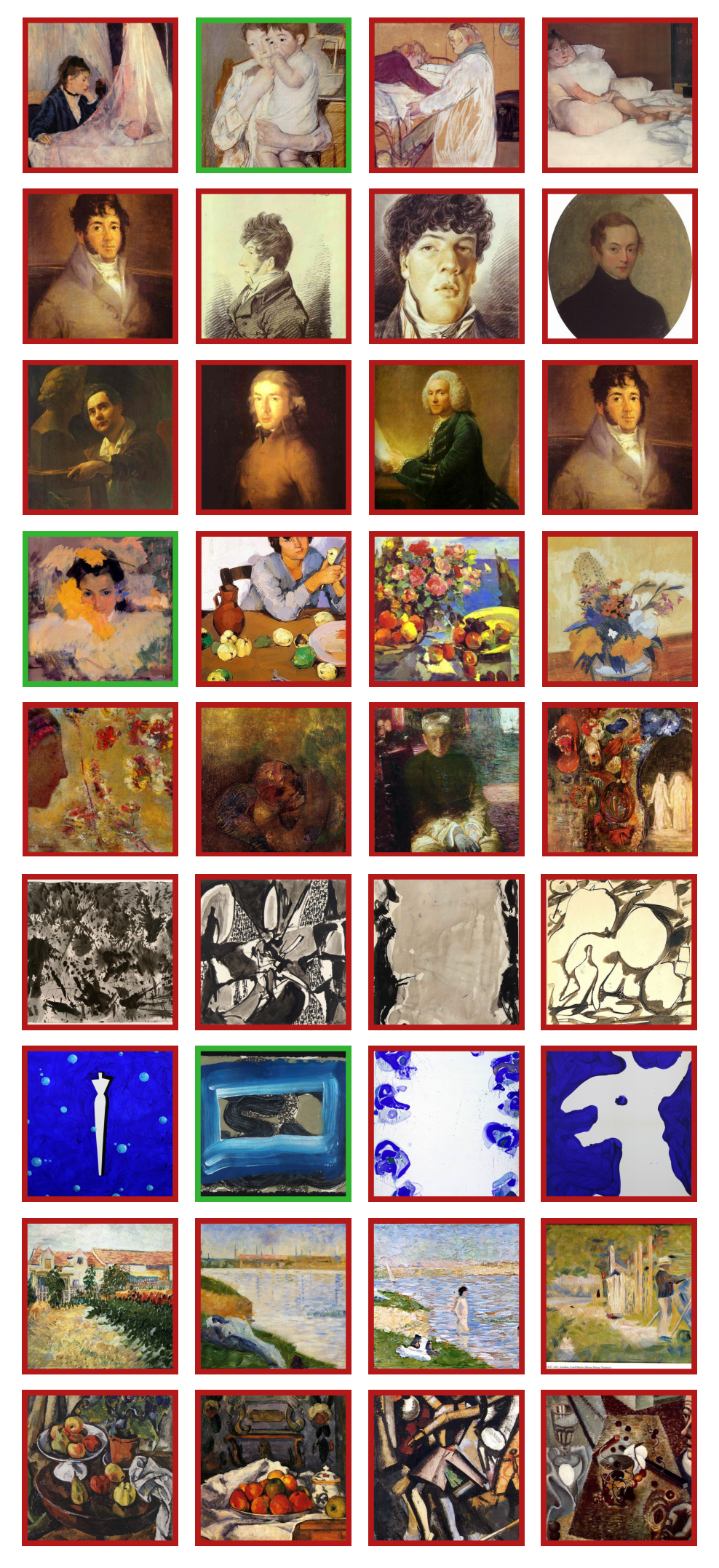}
        \caption{CSD \cite{somepalli2024measuring}}
        
    \end{subfigure}
    \caption{Image Retrieval Results on WikiArt Dataset for \ours, with images ranked highest to lowest from left to right compared with CSD. Green colors indicate \textcolor{correct}{correct} and red for \textcolor{wrong}{incorrect} retrievals.}
    \label{fig:wiki_comp_1}
\end{figure*}

\begin{figure*}[t!]
    \centering
    \begin{subfigure}[b]{0.10\textwidth}
        \centering
        \includegraphics[width=1.13\textwidth]{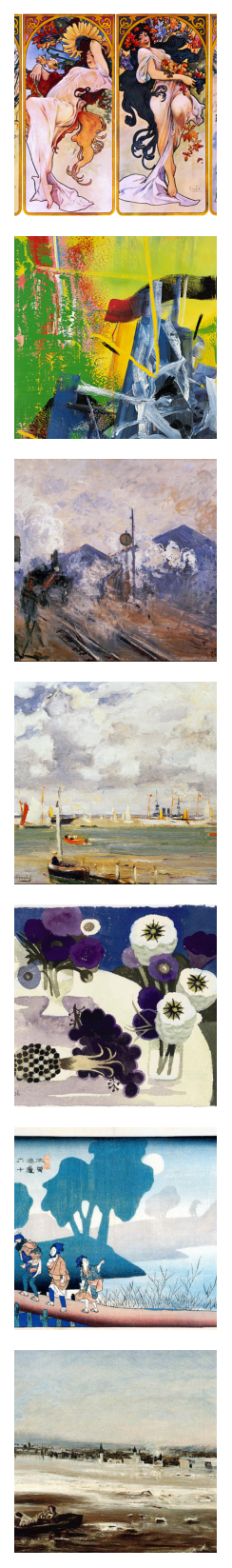}
        \caption{Queries}
        
    \end{subfigure}
    \hfill
    \begin{subfigure}[b]{0.89\textwidth}
        \centering
        \includegraphics[width=1\textwidth]{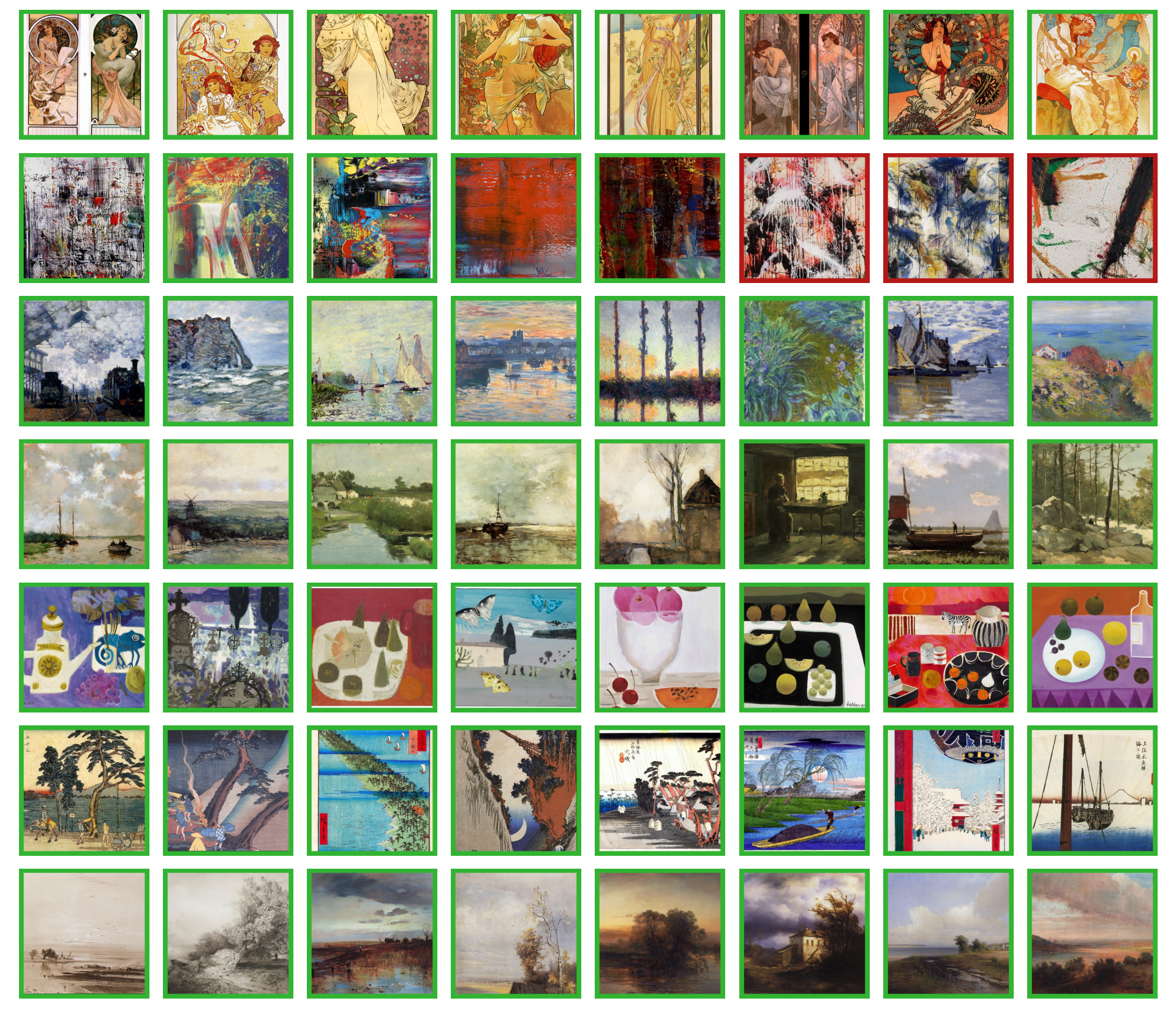}
        \caption{\ours (ours)}
        
    \end{subfigure}
    \caption{Additional Image Retrieval on WikiArt Dataset for \ours with images ranked highest to lowest from left to right. Green colors indicate \textcolor{correct}{correct} and red for \textcolor{wrong}{incorrect} retrievals.}
    \label{fig:wikiart_extra}
\end{figure*}

\begin{figure*}[t!]
    \centering
    \begin{subfigure}[b]{0.10\textwidth}
        \centering
        \includegraphics[width=1.13\textwidth]{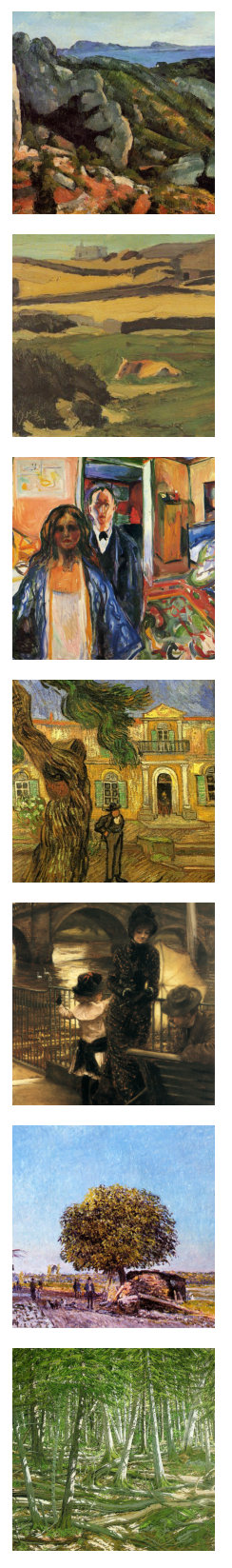}
        \caption{Queries}
        
    \end{subfigure}
    \hfill
    \begin{subfigure}[b]{0.89\textwidth}
        \centering
        \includegraphics[width=1\textwidth]{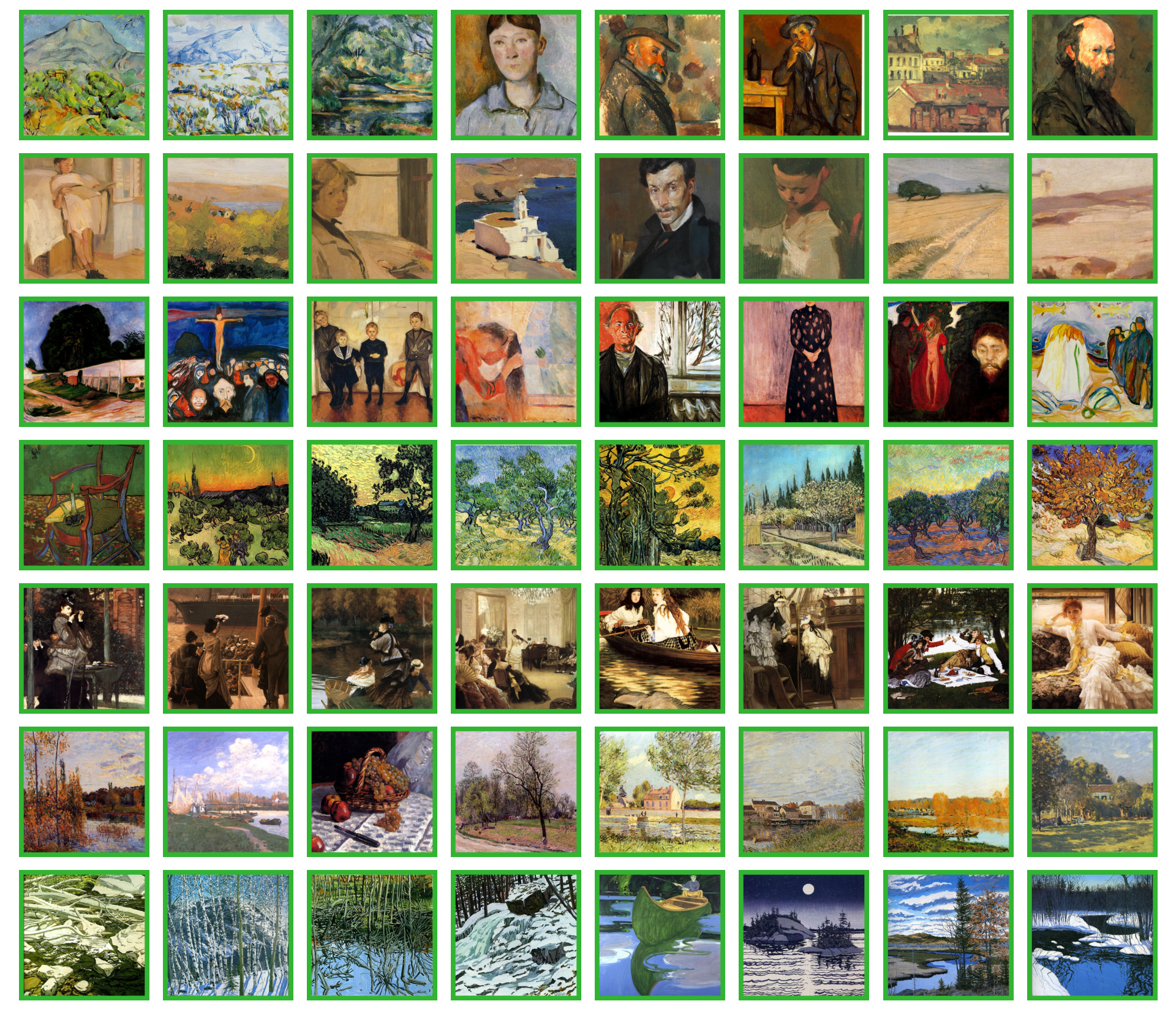}
        \caption{\ours (ours)}
        
    \end{subfigure}
    \caption{Additional Image Retrieval on WikiArt Dataset for \ours with images ranked highest to lowest from left to right. Green colors indicate \textcolor{correct}{correct} and red for \textcolor{wrong}{incorrect} retrievals.}
    \label{fig:wikiart_extra_2}
\end{figure*}

\begin{figure*}[t!]
    \centering
    \begin{subfigure}[b]{0.105\textwidth}
        \centering
        \includegraphics[width=1\textwidth]{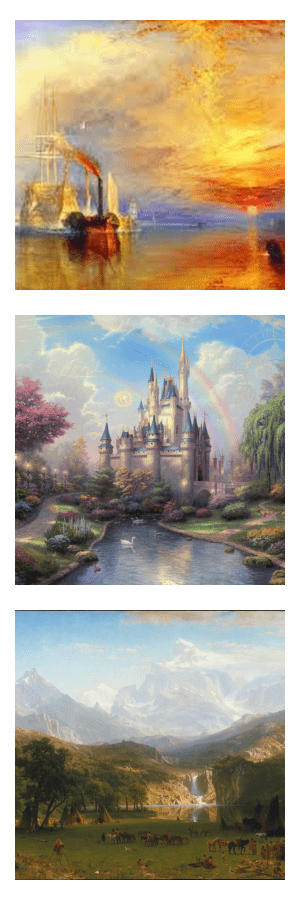}
        \caption{Queries}
        
    \end{subfigure}
    \hfill
    \begin{subfigure}[b]{0.42\textwidth}
        \centering
        \includegraphics[width=1\textwidth]{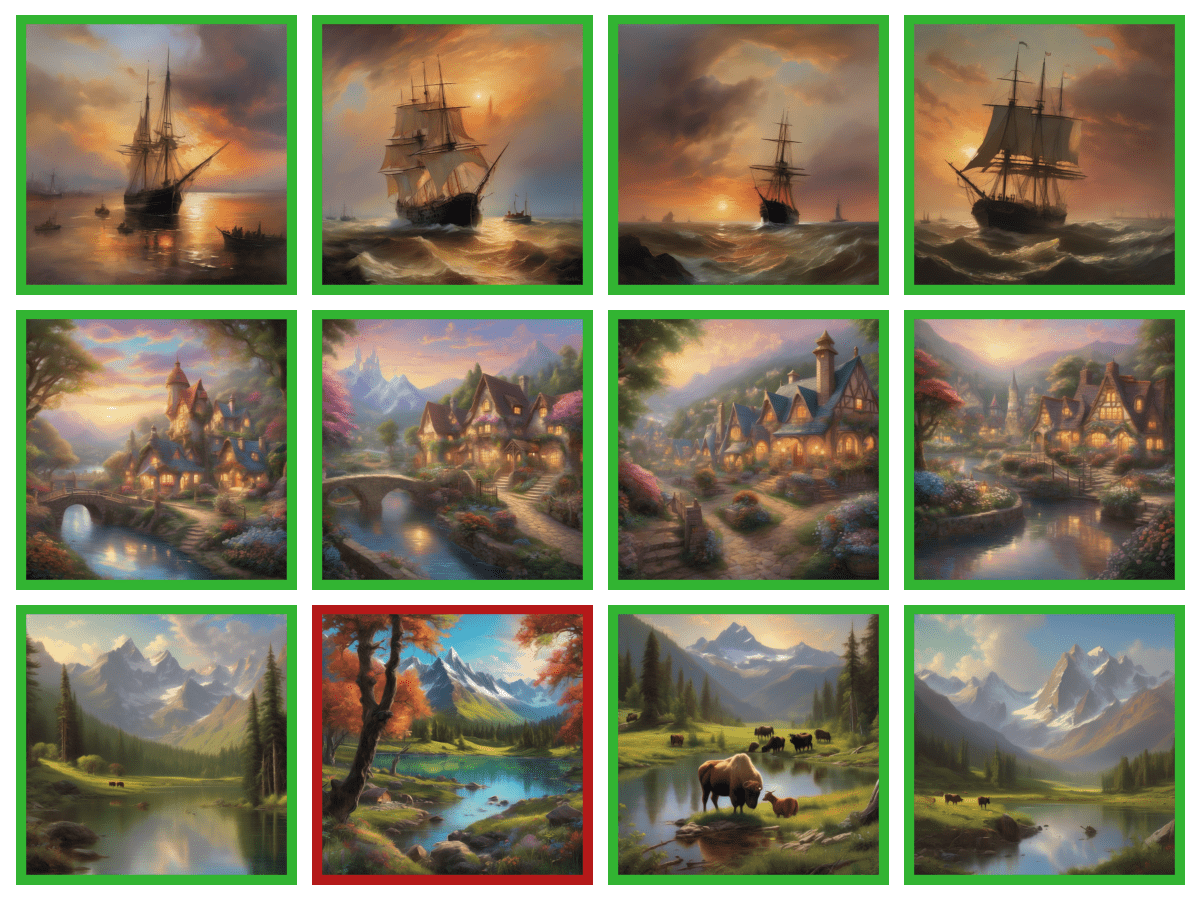}
        \caption{\ours (ours)}
        
    \end{subfigure}
    \hfill
    \begin{subfigure}[b]{0.42\textwidth}
        \centering
        \includegraphics[width=1\textwidth]{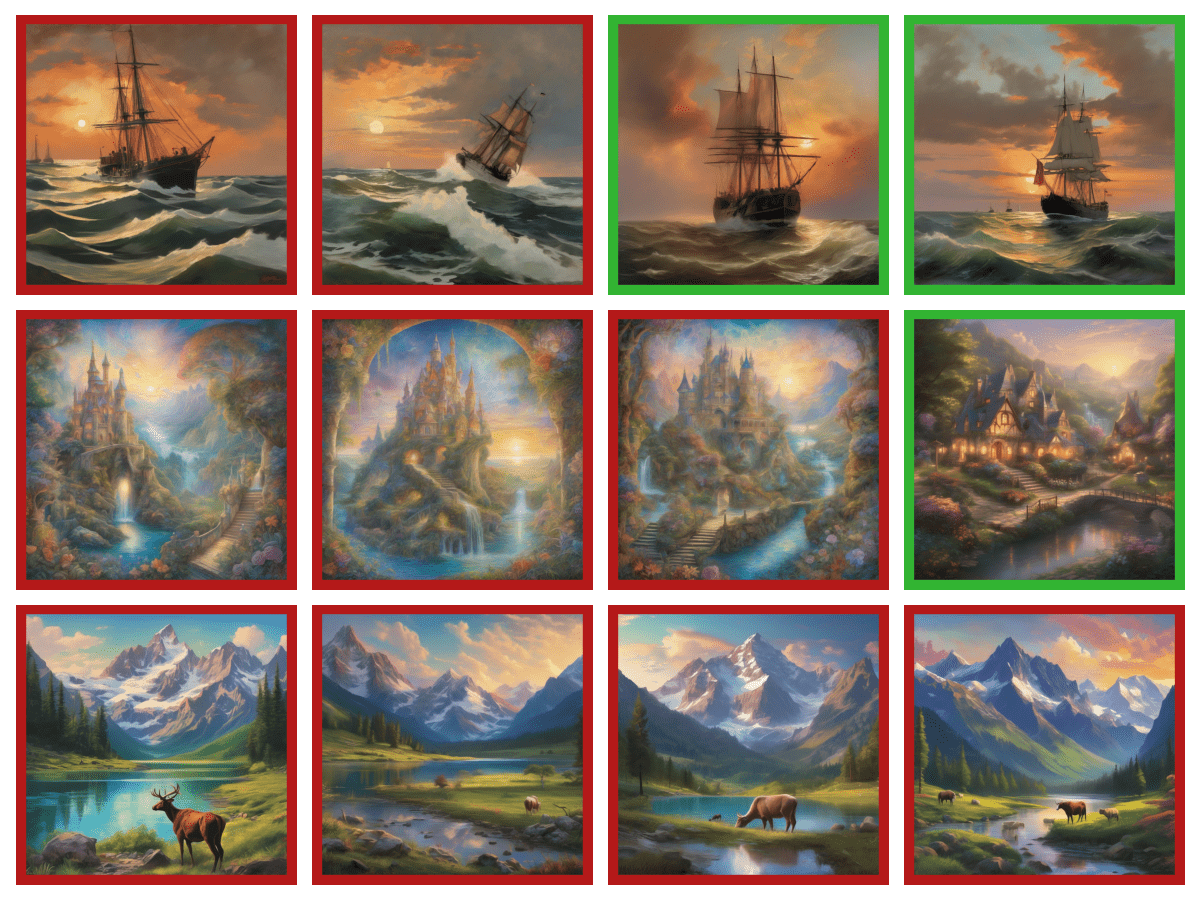}
        \caption{CSD \cite{somepalli2024measuring}}
        
    \end{subfigure}
    \caption{Additional Image Retrieval Results for Style-based Evaluation on \ourd Dataset, with images ranked highest to lowest from left to right. The images are filtered to a fixed semantic. Green colors indicate \textcolor{correct}{correct} and red for \textcolor{wrong}{incorrect} retrievals.}
    \label{fig:synth_res_style_supp}
\end{figure*}

\begin{figure*}[t!]
    \centering
    \begin{subfigure}[b]{0.105\textwidth}
        \centering
        \includegraphics[width=1\textwidth]{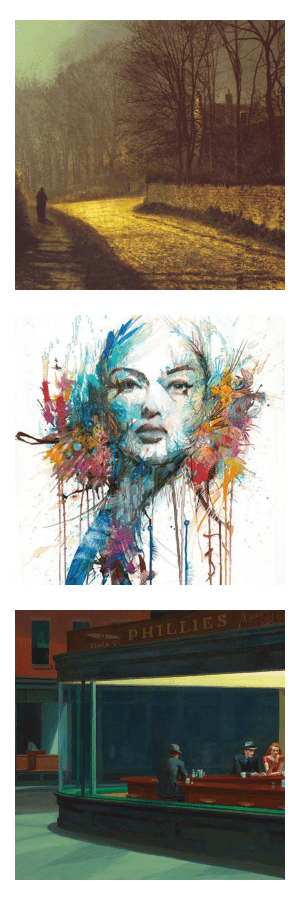}
        \caption{Queries}
        
    \end{subfigure}
    \hfill
    \begin{subfigure}[b]{0.42\textwidth}
        \centering
        \includegraphics[width=1\textwidth]{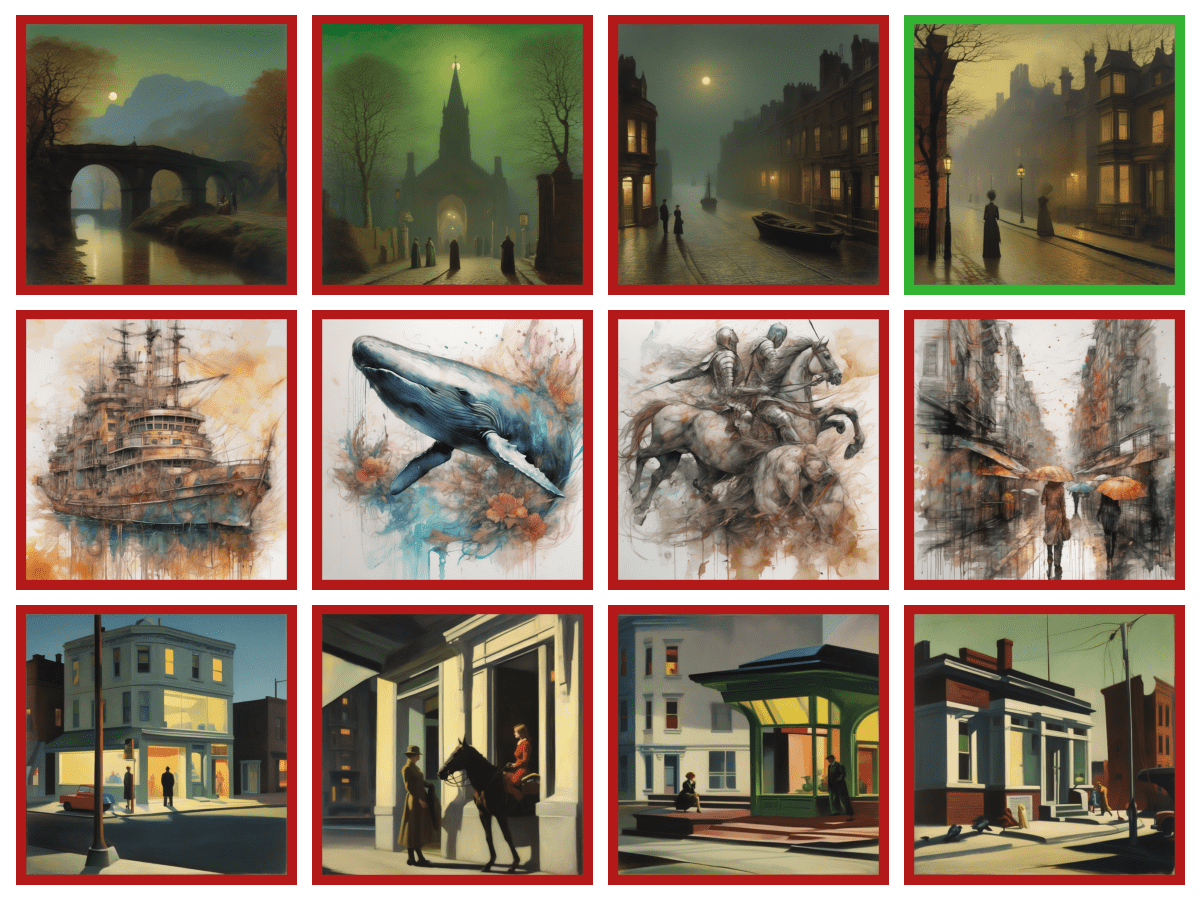}
        \caption{\ours (ours)}
        
    \end{subfigure}
    \hfill
    \begin{subfigure}[b]{0.42\textwidth}
        \centering
        \includegraphics[width=1\textwidth]{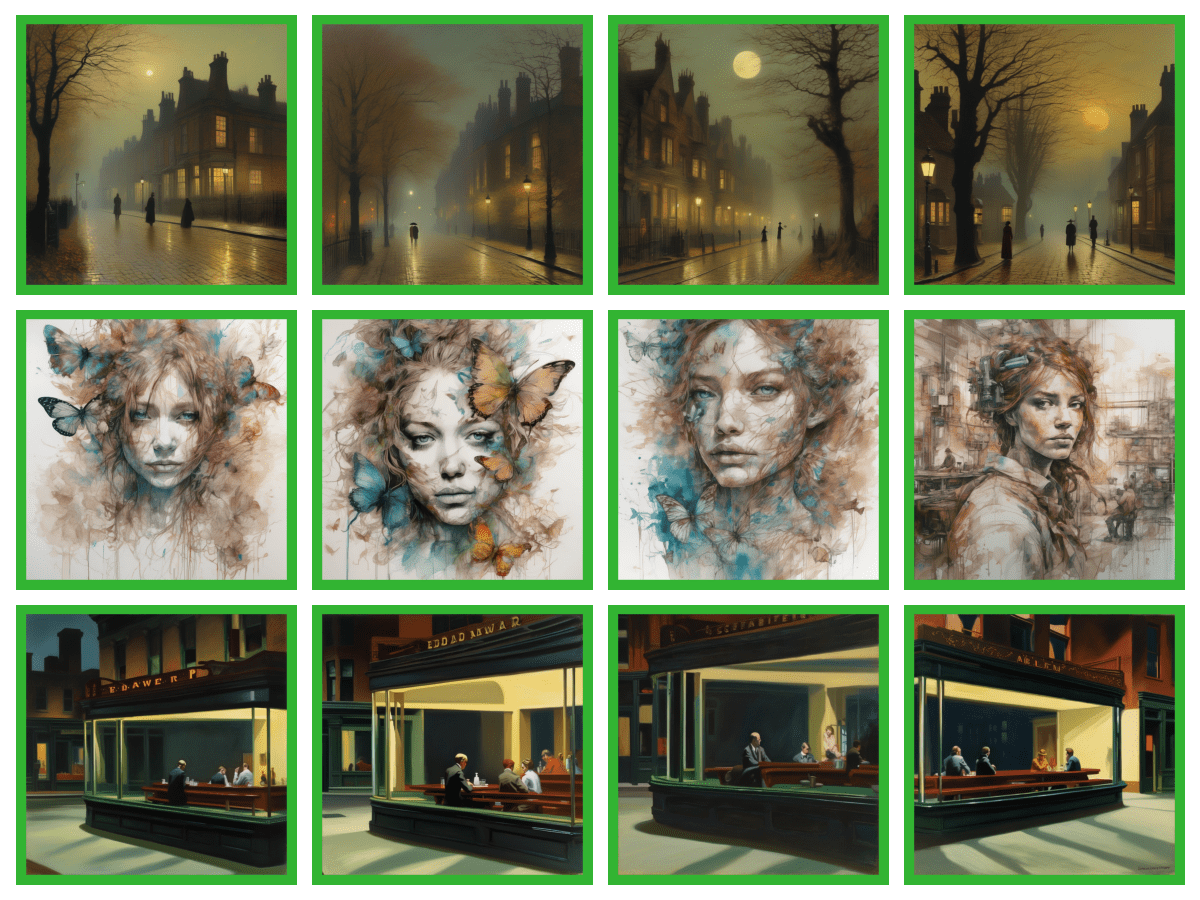}
        \caption{CSD \cite{somepalli2024measuring}}
        
    \end{subfigure}
    \caption{Additional Image Retrieval Results for Semantic-based Evaluation on \ourd Dataset, with images ranked highest to lowest from left to right. Green colors indicate \textcolor{correct}{correct} and red for \textcolor{wrong}{incorrect} retrievals.}
    \label{fig:synth_res_sem_supp}
\end{figure*}

Figs.~\ref{fig:wikiart_extra} and~\ref{fig:wikiart_extra_2} show retrieval results of the proposed \ours method on WikiArt and Fig~\ref{fig:wiki_comp_1} compares our retrievals with CSD. Remarkably, \ours can identify images of the same style despite a great diversity of content. Note that the images retrieved rarely present images of similar content, indicating the robustness of the proposed approach in focusing on styles. Furthermore, we show more retrieval results in Figs.~\ref{fig:synth_res_style_supp} and~\ref{fig:synth_res_sem_supp} for the \ourd Dataset, where our method works effectively and achieves high retrieval accuracy over state-of-the-art method (CSD).

% \begin{figure*}[t!]
%     \centering
%     \begin{subfigure}[b]{0.128\textwidth}
%         \centering
%         \includegraphics[width=1\textwidth]{sec/figures/wikiart/ref_1_supp.png}
%         \caption{Queries}
        
%     \end{subfigure}
%     \hfill
%     \begin{subfigure}[b]{0.42\textwidth}
%         \centering
%         \includegraphics[width=1\textwidth]{sec/figures/wikiart/ours_1_supp.png}
%         \caption{\ours (ours)}
        
%     \end{subfigure}
%     \hfill
%     \begin{subfigure}[b]{0.42\textwidth}
%         \centering
%         \includegraphics[width=1\textwidth]{sec/figures/wikiart/csd_1_supp.png}
%         \caption{CSD \cite{somepalli2024measuring}}
        
%     \end{subfigure}
%     \caption{Image Retrieval Results on WikiArt Dataset for \ours, with images ranked highest to lowest from left to right compared with CSD. Green colors indicate \textcolor{correct}{correct} and red for \textcolor{wrong}{incorrect} retrievals.}
%     \label{fig:wiki_comp_2}
% \end{figure*}

% \begin{figure*}[t!]
%     \centering
%     \begin{subfigure}[b]{0.12\textwidth}
%         \centering
%         \includegraphics[width=1\textwidth]{sec/figures/synth/ref_cmp_sup.png}
%         \caption{Queries}
        
%     \end{subfigure}
%     \begin{subfigure}[b]{0.6225\textwidth}
%         \centering
%         \includegraphics[width=1.13\textwidth]{sec/figures/synth/ours_sup.png}
%         \caption{\ours (ours)}
        
%     \end{subfigure}
%     \caption{Additional Image Retrieval on \ourd Dataset for \ours with images ranked highest to lowest from left to right. Green colors indicate \textcolor{correct}{correct} and red for \textcolor{wrong}{incorrect} retrievals.}
%     \label{fig:SYNTH_extra}
% \end{figure*}

% \section{Rationale}
% \label{sec:rationale}
% % 
% Having the supplementary compiled together with the main paper means that:
% % 
% \begin{itemize}
% \item The supplementary can back-reference sections of the main paper, for example, we can refer to \cref{sec:intro};
% \item The main paper can forward reference sub-sections within the supplementary explicitly (e.g. referring to a particular experiment); 
% \item When submitted to arXiv, the supplementary will already included at the end of the paper.
% \end{itemize}
% % 
% To split the supplementary pages from the main paper, you can use \href{https://support.apple.com/en-ca/guide/preview/prvw11793/mac#:~:text=Delete%20a%20page%20from%20a,or%20choose%20Edit%20%3E%20Delete).}{Preview (on macOS)}, \href{https://www.adobe.com/acrobat/how-to/delete-pages-from-pdf.html#:~:text=Choose%20%E2%80%9CTools%E2%80%9D%20%3E%20%E2%80%9COrganize,or%20pages%20from%20the%20file.}{Adobe Acrobat} (on all OSs), as well as \href{https://superuser.com/questions/517986/is-it-possible-to-delete-some-pages-of-a-pdf-document}{command line tools}.

\end{document}